\documentclass[journal]{IEEEtran}
\usepackage{siunitx} 
\usepackage[ruled]{algorithm2e}
\usepackage{amssymb}
\usepackage{stackengine}
\usepackage{amsmath}
\usepackage{subfigure}
\usepackage{amsthm}
\usepackage{multirow}
\usepackage{diagbox}
\usepackage[colorlinks,linkcolor=black, anchorcolor=black,citecolor=black]{hyperref}
\hypersetup{urlcolor=black}
\usepackage{arydshln}
\usepackage{etoolbox}
\usepackage{enumerate}
\usepackage{booktabs}
\usepackage[justification=centering,font={small}]{caption}

\usepackage{cite} 

\usepackage{amsfonts, amsmath, mathrsfs,  amssymb, dsfont, mathtools, slashbox, pifont, amsbsy, graphicx,  yfonts, bbm, bm}
\usepackage[all,dvips,arc,curve,color,frame]{xy}

\newcommand\prefixtext[1]{%
  \ifvmode\else\\\@empty\fi
  \noalign{%
    \penalty0%
    \vbox{\mathstrut}%
    \penalty10000%
    \vskip-\baselineskip
    \penalty10000%
    \vbox to 0pt{%
      \normalbaselines
      \ifdim\linewidth=\columnwidth
      \else
        \parshape\@ne
        \@totalleftmargin\linewidth
      \fi
      \vss
      \noindent#1\par}%
      \penalty10000%
      \vskip-\baselineskip}%
      \penalty10000}

\DeclareMathAlphabet{\mathpzc}{OT1}{pzc}{m}{it}
\newcommand{\comment}[1]{}

\def\({\left(}
\def\){\right)}
\def\[{\left[}
\def\]{\right]}
\def\BEq{\begin{eqnarray}}
\def\EEq{\end{eqnarray}}
\def\BE*{\begin{eqnarray*}}
\def\EE*{\end{eqnarray*}}
\def\BA{\begin{array}}
\def\EA{\end{array}}

\def\0{\mathbf{0}}
\def\1{\mathbf{1}}

\def\G{\mathbf{G}}

\def\bR{\mathbb{R}}

\def\S{\mathbf{S}}

\def\U{\mathbf{U}}

\def\V{\mathbf{V}}

\def\W{\mathbf{W}}

\def\X{\mathbf{X}}

\def\tY{\bm{\mathcal{Y}}}

\def\tU{\tilde{\U}_s}

\def\tV{\tilde{\V}_s}

\def\tX{\bm{\mathcal{X}}}

\def\tW{\bm{\mathcal{W}}}

\def\sc{\bm{c}}

\def\tU{\bm{\mathcal{U}}}
\def\tV{\bm{\mathcal{V}}}

\def\and{\prefixtext{and}}

\newcommand\cb{\color{black}}
\newcommand\cred{\color{black}}

\newcommand\cbl{\color{black}}

\begin{document}

\title{Joint Matrix Decomposition for Deep Convolutional Neural Networks Compression}

\author{ Shaowu Chen,
              Jiahao Zhou,
              Weize Sun*,
              Lei Huang

\thanks{Shaowu Chen, Jiahao Zhou, Weize Sun and Lei Huang are with the College of Electronics and Information Engineering, Shenzhen University, Shenzhen 518060, Guangdong, China.
(e-mail: shaowu-chen@foxmail.com;
plus\_chou@foxmail.com;
proton198601@hotmail.com;
lhuang8sasp@hotmail.com).
}
\thanks{*Corresponding author:  Weize Sun.}
}
\maketitle

\begin{abstract}
Deep convolutional neural networks (CNNs) with a {\cbl large} number of parameters require intensive computational resources,
and thus are hard to be deployed in resource-constrained platforms. Decomposition-based methods, therefore, have been utilized to compress CNNs in recent years. However, since the compression factor and performance are negatively correlated, the state-of-the-art works either suffer from severe performance degradation or have relatively low compression factors. To overcome this problem, we propose to compress CNNs and alleviate performance degradation via joint matrix decomposition, which is different from existing works that compressed layers separately. The idea is inspired by the fact that there are lots of repeated modules in CNNs. By projecting weights with the same structures into the same subspace, networks can be jointly compressed with larger ranks. In particular, three joint matrix decomposition schemes are developed, and the corresponding optimization approaches based on Singular Value Decomposition are proposed. Extensive experiments are conducted across three challenging compact CNNs for different benchmark data sets to demonstrate the superior performance of our proposed algorithms. As a result, our methods can compress the size of ResNet-34 by $22\times$ with slighter accuracy degradation compared with several state-of-the-art methods.
\end{abstract}


\section{Introduction}
\label{sec:intro}
In recent years, deep neural networks (DNNs), including deep convolutional neural networks (CNNs) and multilayer perceptron (MLPs), have achieved great successes in various areas, \emph{e.g.}, noise reduction, object detection and matrix completion \cite{DBLP:journals/ijon/GuanLXLG20, DBLP:conf/aaai/TaherkhaniKN19, Girshick2014ObjectDetection}. To achieve satisfactory performance, very deep and complicated DNNs with a cumbersome number of parameters and billions of floating point operations (FLOPs) requirements are developed \cite{EfficientNet,VGG,He2016ResNet, GoogleNet}. Although high-performance servers with GPUs can meet the requirements of the DNNs, it is problematic to deploy them on resource-constrained platforms such as embedded or mobile devices \cite{Kim2016Tucker,SqueezeNet}, especially when real-time forward inference is required. To tackle this problem, methods for network compression are developed.

It has been found that the weight matrices of fully connected (FC) layers and weight tensors of convolutional layers are of low rank \cite{Denil2013Predicting}, therefore the redundancy among these layers can be removed via decomposition methods. In \cite{Denil2013Predicting, Denton2014Exploiting, Jaderberg2014Speeding, Tai2016Convolutional}, matrix decomposition alike methods are utilized to compress MLPs and CNNs in a one-shot manner or layer by layer progressively, in which a weight matrix $\W$ is decomposed as $\W\approx\U\V$ where $\textrm{size}(\U)+\textrm{size}(\V)\ll\textrm{size}(\W)$, and relatively small compression factors (CF) are achieved at the cost of a drop in accuracy. Recent works focus more on compressing CNNs, and to avoid unfolding 4-D weight tensors of convolutional layers into 2-D matrices when implementing decomposition, tensor-decomposition-based methods \cite{Novikov2015TT,Kim2016Tucker,CPICLR,Garipov2016TTalike,Wang2018TensorRing} are introduced to compress CNNs. However, CNNs are much more compacted than MLPs because of the properties of parameter sharing, making the compression of CNNs challenging, thus previous works either compress CNNs with a small CF between $2\times$ and $5\times$ or suffer from severe performance degradation. Some methods are proposed to alleviate degradation with a scheme associating properties of low rank and sparsity \cite{Yu2017LowRankSparse,junhao20}, in which $\W\approx\U\V+\S$, where $\S$ is a highly sparse matrix. However, without support from particular libraries, the memory, storage and computation consumption of $\S$ is as large as $\W$ since $\textrm{size}(\S)=\W$, and the {\cbl ``}0{\cbl "} elements as $2^n$-bits numbers consume the same resources as those non-zero ones, not to mention the ones caused by $\U$ and $\V$, making the method less practical.

\begin{figure*}
    \centering
    \includegraphics[width=1\linewidth]{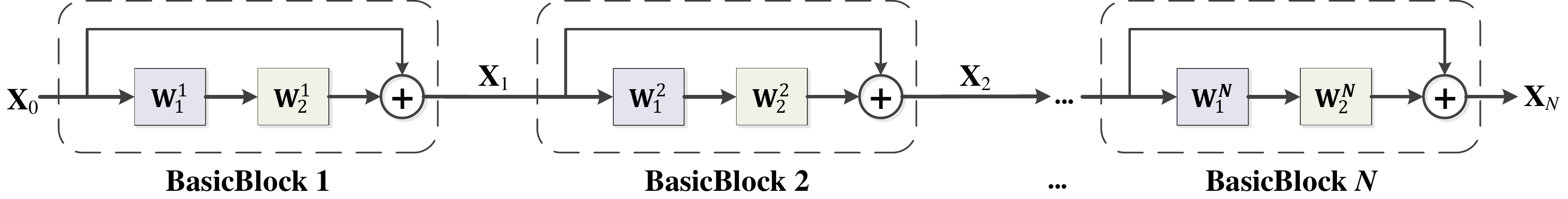}
    \caption{A common network structure in ResNet.
    Here we assume that weights of convolutional layers are 2-D matrices instead of 4-D tensors for the convenience of explanation.}
    \label{fig:RepeatModule}
\end{figure*}

To greatly compress CNNs without severe performance degradation, we propose to compress them via joint matrix decomposition. The main difference between our scheme and the state-of-the-art works is that we jointly decompose layers with relationships instead of compressing them separately. The idea is inspired by a basic observation that the widely used CNNs tend to adapt repeated modules to attain satisfying performance \cite{VGG,He2016ResNet, GoogleNet}, making lots of convolutional layers sharing the same structure. Therefore, by projecting them into the same subspace, weight tensors can share identical factorized matrices, and thus CNNs can be further compressed. Taking part of ResNet shown in Figure \ref{fig:RepeatModule} as an example,  $\W_1^n, n=1,\cdots, N$ have the same structure and are placed in the consistent position of N BasicBlocks, thus might contain similar information and can be jointly decomposed, \textit{i.e.}, $\W_1^n=\G_1^n\Sigma_1^n\V_1=\U_1^n\V_1$ where $\V_1$ is identical for all  $\W_1^n, n=1,\cdots,N$, and  $\U_1^n=\G_1^n\Sigma_1^n$. In this manner, there are only about half of the matrices needed to be stored, \textit{i.e.}, $\U_1^1,\U_1^2,\cdots,\U_1^N$ and $\V_1$ instead of $\U_1^1,\U_1^2,\cdots,\U_1^N$ and $\V_1^1,\V_1^2,\cdots,\V_1^N$, and thus the requirement of storage and memory resources can be further reduced. {\cred With the shared $\V_1$ containing the common information across multiple layers in large CNNs, the joint decomposition methods can retain larger ranks compared with traditional rank-truncated decomposition methods under the same compression degree, and thus alleviate performance degradation in the compressed models.}
After decomposing $\W_1^n, n=1,\cdots,N$, the similar procedure can be implemented on $\W_2^n, n=1,\cdots,N$ as well. In our previous work \cite{JSTSP}, the relationship of layers was also taken into account but with a clearly different idea, which considered a weight as the summation of an independent component and a shared one, and Tucker Decomposition \cite{Kim2016Tucker} or Tensor Train Decomposition \cite{TensorTrainDecompositon} was utilized to factorized them, \textit{i.e.}, $\W^n_m=\W_{m,i}^n+\W_{m,s}=\G_{m,i,1}^n*\G_{m,i,2}^n*\G_{m,i,3}^n+\G_{m,s,1}*\G_{m,s,2}*\G_{m,s,3}$, for $n=1,\cdots,N$, while in this paper we directly project weights into the same subspace via joint matrix decomposition.

In the above illustration, for convenience, the weight $\W$ in a convolutional layer is supposed to be two-dimensional, while they are 4-D tensors in reality. To implement joint matrix decomposition, we firstly unfold 4-D weight tensors into 2-D weight matrices following \cite{Tai2016Convolutional}, and then introduce a novel Joint Singular Value Decomposition (JSVD) method to decompose networks jointly. Two JSVD algorithms for network compression are proposed,
referred to as left shared JSVD (\textbf{LJSVD}) and right shared JSVD (\textbf{RJSVD}), respectively, according to the relative position of the identical factorized matrix. Besides, by combining LJSVD and RJSVD, another algorithm named Binary JSVD (\textbf{Bi-JSVD}) is also proposed, which can be considered as the generalized form of LJSVD and RJSVD. These algorithms can decompose pre-trained CNNs following the {\cbl ``}pre-train$\to$decompose$\to$fine-tune{\cbl''} pipeline
and train compressed CNNs from scratch by retaining the decomposed structures but randomly initializing weights. Furthermore,
as a convolutional layer can be divided into two consecutive slimmer ones with less complexity by matrix decomposition, our methods can not only compress networks but also accelerate them without support from customized libraries and hardware.

The remainder of this paper is organized as follows. In Section \ref{sec:relatedworks}, some of the related works are reviewed. In Section \ref{sec:methodology}, necessary notations and definitions are introduced first, and then the proposed joint matrix decomposition methods for network compression are developed.
Section \ref{sec:experiments} presents extensive experiments evaluating the proposed methods, and the results are discussed in detail. Finally, a brief conclusion and future work are given in Section \ref{sec:conclusion}.

\section{Related Works}
\label{sec:relatedworks}
{\cb
The methods for neural network compression  can be roughly divided into four categories, namely, pruning, quantization, knowledge distillation and decomposition.
Furthermore, the decomposition methods can be further divided into matrix-based and tensor-based ones.
Note that the pruning, quantization and knowledge distillation approaches are orthogonal to decomposition methods, and thus can be combined to achieve better performance in some cases.}

{\cb
\paragraph{Pruning}
Pruning methods evaluate the importance of neurons or filters and eliminate those with the lowest scores. Its origins can date back to the 1990s \cite{lecun1990optimal}, which used second-order derivative information to prune neurons that bring minor changes to the loss function. In \cite{han2015learning,LotteryTicket}, weights with magnitudes lower than a threshold are pruned iteratively, while the works in \cite{junhao20,Chen2021MOEA} pruned CNNs using evolutionary computing methods. The aforementioned methods pruned the elements in weight tensors by setting them to zero, resulting in unstructured sparsity that is less effective for compression and acceleration.
In contrast, the works in \cite{PPCA,GeometricFilterPruning} directly removed redundant filters and thus can effectively compress CNNs.

\paragraph{Quantization}
Some researchers compress and accelerate CNNs using low-precision and fixed-point number representations for weights. The works in \cite{JacobInteger} and \cite{XNOR} successfully utilized binary and integer numbers, respectively, to reduce the sizes and expensive floating point operations of CNNs. These works used quantization techniques in the training stage, while some post-quantization approaches \cite{HanDeepCompression,WangPostQuan} trained full-precision CNNs at first and then quantized them to obtain low-precision results. Similar to our methods, most post-training quantization {\cbl methods require} fine-tuning to alleviate performance degradation.

\paragraph{Knowledge distillation}
In general, larger networks contain more knowledge and thus outperform smaller networks. Therefore, knowledge distillation aims at transferring the knowledge of large pre-trained teacher-CNNs to small student-CNNs \cite{hinton2015distilling}. After transferring,
the large CNNs are discarded while the smaller teacher-CNNs with similar performance are utilized. Hinton \textit{et al.} \cite{hinton2015distilling} implemented knowledge distillation by reducing the differentiation between softmax output of teacher-CNNs and student-CNNs, and researchers extended knowledge distillation by matching other statistics, such as intermediate feature maps \cite{Romero2015KD,Chen2016KD,Li2020KD} and gradient \cite{Srinivas2018KD}. The idea of knowledge distillation can also be combined for our methods to help fine-tune the compressed CNNs and further improve performance.
}

\paragraph{Matrix decomposition}
In \cite{Denil2013Predicting}, weight tensors of convolutional layers were considered of low rank. Therefore, they were unfolded to 2-D matrices and then compressed via low-rank decomposition, in which the reconstruction ICA \cite{ICA} and ridge regression with the squared exponential kernel were used to predict parameters. Inspired by this,
the work in \cite{Denton2014Exploiting} combined SVD and several clustering schemes to achieve a $3.9\times$ weight reduction for a single convolutional layer. In \cite{Tai2016Convolutional}, CNNs were compressed by removing spatial or channel redundancy via low rank regularization, and a $5\times$ compression on AlexNet was achieved \cite{Tai2016Convolutional}. With the idea similar to knowledge distilling, some researchers initially trained an over-parameter network and then compressed it via matrix decomposition to meet budget requirements, thus making the compressed network outperform the one trained from scratch directly with the same sizes \cite{Sun2017SVDTDNN}. The works in \cite{Yu2017LowRankSparse, junhao20} combined the property of sparsity with the low rank one to obtain higher compression factors, but need supports from customized libraries to implement storage compression and computation acceleration. In \cite{SparseLowRank}, the sparsity was embedded in the factorized low rank matrices, which can be considered as the combination of unstructured pruning and matrix decomposition.

\paragraph{Tensor decomposition}
Since the weight of a convolution layer is a 4-D tensor, tensor decomposition methods were applied to compress neural networks naturally to achieve higher compression factors. Some compact modules can be regarded as derivations from tensor decomposition as well \cite{Decoupling}, such as the depthwise separable convolution in MobileNet \cite{MobileNets}. In \cite{Garipov2016TTalike}, Tensor Train Decomposition (TTD) was extended from MLPs \cite{Novikov2015TT} to compress CNNs. To achieve a higher degree of reduction in sizes, weight tensors in \cite{Garipov2016TTalike} were folded to higher dimensional tensors and then decomposed via TTD. As a result, a CNN were compressed by $4.02\times$ with a $2\%$ loss of accuracy. In \cite{Kim2016Tucker}, spatial dimensions of a weight tensor were merged to form a 3-D tensor, and then Tucker Decomposition was utilized to divide the original convolutional layer into three layers with less complexity. Note that in this case, the Tucker Decomposition is equal to TTD. To alleviate performance degradation caused by Tucker, the work in \cite{JSTSP} considered a weight tensor as the summation of the independent and shared components, and applied Tucker Decomposition to each of them, respectively.
In \cite{wu2020hybrid}, Hybrid Tensor decomposition was utilized to achieve $4.29\times$ compression across a CNN on CIFAR-10 at the cost of $5.18\%$ loss in the accuracy. At the same time, TTD \cite{Garipov2016TTalike} only caused $1.27\times$ loss, indicating that TTD is more suitable for compressing convolution layers.

\section{Methodology}
\label{sec:methodology}
In this section, some necessary notations and definitions are introduced first, and then three joint compression algorithms, RJSVD, LJSVD and Bi-JSVD, are proposed.

\subsection{Preliminaries}
\label{Preliminaries}

\textbf{Notations}.
The notations and symbols used in this paper are introduced as follows. Scalars, vectors, matrices (2-D), and tensors (with more than two dimensions) are denoted by italic, bold lowercase, bold uppercase, and bold calligraphic symbols, respectively. Following the conventions of Tensorflow,
we represent an input tensor of one convolution layer as $\tX \in \mathbb{R}^{H_1 \times W_1 \times I}$, the output as $\tY \in \mathbb{R}^{H_2 \times W_2 \times O }$, and the corresponding weight tensor as $\tW \in \bR^{F_1 \times F_2 \times I \times O}$, where $H_1,W_1,H_2,W_2$ are spatial dimensions, $F_1 \times F_2$ is the size of a filter, while $I$ and $O$ are the input and output depths, respectively.
Sometimes we would put the sizes of a tensor or matrix in its subscript to clarify the dimensions, such as $\W_{I \times O}$ means $\W \in \mathbb{R}^{I \times O}$.

To  decompose a group of layers jointly, we further denote weights tensors or matrices with subscripts and superscripts such as $\tW^{n}_m$, for $n=1,\cdots,N$ and $m=1,\cdots,M$,
where $m$ distinguishes different groups and $n$ distinguishes different elements inside a group. In other words, $\tW^{n}_m$, $n=1,\cdots,N$ under the same $m$ would be jointly decomposed. Figure \ref{fig:RepeatModule} is a concrete example with $M=2$, and $\W_1^n$, $n=1,2,\cdots,N$ is a group of weights that would be jointly decomposed, so as $\W_2^n$, $n=1,2,\cdots,N$.

\textbf{General unfolding}.
When applying matrix decomposition to compress convolutional layers, the first step is to unfold weight tensors $\tW \in \bR^{F_1\times F_2 \times I \times O}$ into 2-D matrices. Since the kernel sizes $F_1$ and $F_2$ are usually small values such as 3 or 5, following \cite{Tai2016Convolutional}, we merge the first and third dimensions and then the second and fourth ones.
We refer to this operation as general unfolding and denote it by $\textbf{Unfold}(\cdot)$, which swaps the first and third axes and then merges the first two and the last dimensions, respectively, to produce $\textrm{Unfold}(\tW)=\W \in \mathbb{R}^{F_1I \times F_2O}$.

\textbf{General folding}.
We call the opposite operation of general unfolding as general folding, denoted as $\textbf{Fold}(\cdot)$. By performing general folding on  $\W \in \mathbb{R}^{F_1I \times F_2O}$, there would be $\textrm{Fold}(\W)=\tW \in \mathbb{R}^{F_1 \times F_2 \times I \times O}$. For ease of presentation, in the remainder of this article, \textbf{notations $\tW$ and $\W$ would represent the general unfold weight tensor and the corresponding general folded matrix, respectively}, unless there are particular statements.

\textbf{Truncated rank-$r$ SVD}.
The $\textrm{SVD}_r(\cdot)$ represents the truncated rank-$r$ SVD for networks compression, where $r$ is the truncated rank.
Note that $r$ is a user-defined hyper-parameter that decides the number of singular values and vectors in the compressed model.
By performing truncated rank-$r$ SVD, we have
$\W_{F_1I \times F_2O}\approx \G_{F_1I\times r}\Sigma_{r\times r}\V_{r \times F_2O}=\U_{F_1I\times r}\V_{r \times F_2O}$,
where
$\Sigma$ is a diagonal matrix {\cbl consisting} of singular values in descending order,
$\G$, $\V$ are orthogonal factorized matrices,
$\U=\G\Sigma$,
and $r \le \textrm{min}\{F_1I,F_2O\}$.
After the decomposition, it is $\U$ and $\V$ that are stored instead of $\W$ or $\tW$, therefore the network is compressed by a factor of $\frac{F_1F_2IO}{r(F_1I+F_2O)}\times$  where $r < \frac{F_1F_2IO}{(F_1I+F_2O)}$.

\subsection{Proposed Algorithms}
\subsubsection{Right Shared Joint SVD (RJSVD)}
For $\tW^{n}_m\in \bR^{F_1\times F_2 \times I\times O}$, $n=1,\cdots,N$ \textbf{under the same} $\textbf{m}$,
we {\cbl expect} to decompose these $N$ weight tensors jointly as follows:
\begin{align}
\textrm{Unfold}(\tW^n_m)=\W^n_m\approx\G_m^n\Sigma_m^n\V_{m}=\U^n_m\V_{m},
\label{formulation1:RJSVD}
\end{align}
where $\V_{m}\in \bR^{r_m^r\times F_2O}$ is identical and shared for this group of $\tW^n_m, n=1,\cdots,N$,
while $\U^n_m\in \bR^{F_1I \times r_m^r}=\G_m^n\Sigma_m^n$ are different,
and the $r_m^r$ here is the truncated rank.
In this manner,
the sub-network is compressed by a factor of $\frac{F_1F_2ION}{r_m^r(F_1IN+F_2O)}\times$.

The optimization problem for (\ref{formulation1:RJSVD}) can be formulated as:
\begin{align}
\min_{\U^n_m,\V_m}
||\W^n_m - \U^n_m\V_m||_2^2  \label{optimization1} \\
s.t. \quad \textrm{rank}(\U^n_m\V_m)\le r_m^r \nonumber\\
\textrm{for} \quad n=1,2,\cdots,N. \nonumber
\end{align}
Since it is difficult to optimize the rank,
we take $r_m^r$ as a given parameter,
thus (\ref{optimization1}) can be rewritten as
\begin{align}
\min_{\{\U^n_m\}^{N}_{n=1},\V_m}
\frac{1}{N}\sum_{n=1}^{N}||\W^n_m - \U^n_m\V_m||_2^2,  \label{optimization2}
\end{align}
which can be solved by randomly initializing $\U_m^n$ and $\V_m$ at first and then update them alternatively as follows:

\noindent(1) Given $\{\U^n_m\}_{n=1}^N$,

update $\V_m=\frac{1}{N}\sum_{n=1}^{N}(\U_m^{n\textrm{T}}\U^n_m)^{-1}\U_m^{n\textrm{T}}\W^n_m$.

\noindent(2) Given $\V_m$,

update $\U^n_m=\W^n_m\V_m^\textrm{T}(\V_m\V_m^\textrm{T})^{-1}$,
for $n=1,2,\cdots,N$.

Alternatively,
we could solve it by performing truncated rank-$r_m^r$ SVD on the matrix produced by stacking $\W_m^n, n=1,2,\cdots, N$ vertically:
\begin{align}
\begin{bmatrix}
    \U^1_m\\
    \U^2_m\\
    \cdots\\
    \U^N_m\\
     \end{bmatrix}
, \V_m
=\textrm{SVD}_{r_m^r}(
\begin{bmatrix}
    \W^1_m\\
    \W^2_m\\
    \cdots\\
    \W^N_m\\
     \end{bmatrix}).
     \label{rightshared}
\end{align}
The $\V_m$ here is the right singular matrix that is shared and identical for $\W_m^n, n=1,2,\cdots, N$,
therefore, we refer to the algorithm proposed for compressing CNNs based on (\ref{rightshared}) as \textbf{Right Shared Joint SVD} (\textbf{RJSVD}).
\begin{figure}
  \centering
  \includegraphics[width=1\linewidth]{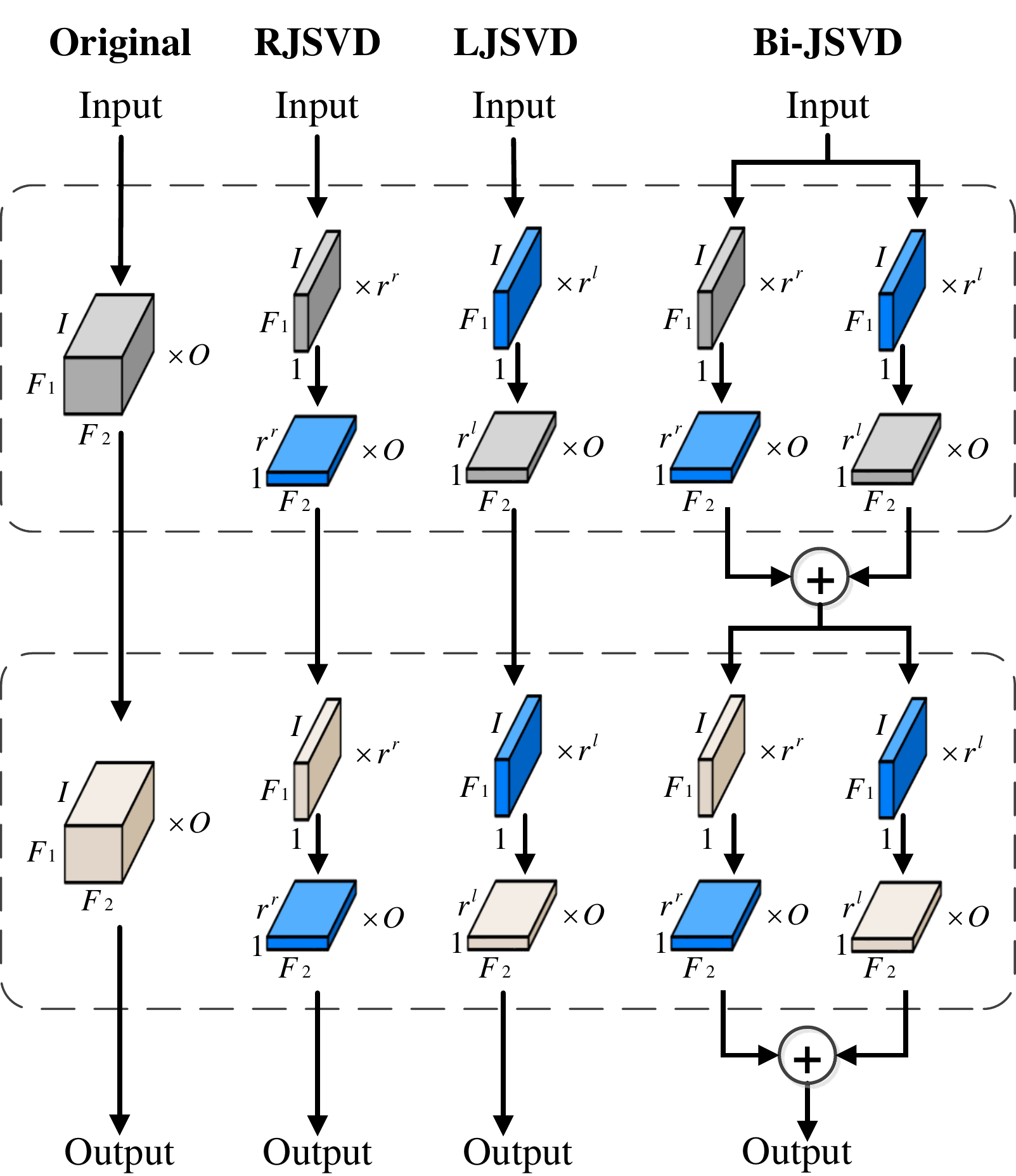}
  \caption{Network structures of the original and the compressed networks with $N=2$.
  RJSVD and LJSVD divide each convolution layer into two consecutive slimmer ones jointly with one half shared (the blue tensors),
   and Bi-JSVD can be seen as a combination of RJSVD and LJSVD.}
\label{fig:net}
\end{figure}
This algorithm can also accelerate the forward inference of CNNs,
because the matrix decomposition shown above actually decomposes a single convolution layer into two successive ones with less complexity,
whose weights are a ``vertical'' tensor $\tU^n_m=\textrm{fold}(\U^n_m) \in \bR^{F_1\times1\times I\times r_m^r}$
and a ``horizontal'' one $\tV_m=\textrm{fold}(\V_m) \in \bR^{1\times F_2 \times r_m^r \times O}$,
respectively,
as shown in Figure \ref{fig:net}
and proved as follows:
\begin{proof}[\quad Proof]
{\cbl Here we} temporarily {\cbl ignore} superscripts and subscripts {\cbl for simplicity}.
{\cbl Suppose that} there is a convolution layer with an input $\tX\in \bR^{F_1 \times F_2 \times I \times O}$ and
a weight tensor $\tW \in \bR^{F_1 \times F_2 \times I\times O}$ whose general unfolding matrix $\W$ can be {\cbl decomposed} into 2 matrices $\U \in \bR^{F_1I\times r}$ and $\V \in \bR^{r\times F_2O}$,
{\cbl then} the convolutional output under settings of $[1,1]$ strides and ``SAME'' padding is $\tY=\textrm{Conv}(\tW,\tX) \in \bR^{1\times1\times O}$.
Defining
$\textrm{vec}(\cdot)$ as vectorization operator,
$\X=\textrm{unfold}(\tX)\in \bR^{F_1I\times F_2O}$
and
$\W_{(:,i,F_2)}=\W[:,F_2*(i-1):F_2*i]$,
\textit{i.e.},
the i-th slice along the second axe with a length of $F_2$,
we have
\begin{align}
\label{proof}
\textrm{vec}(\tY)\nonumber
&=\textrm{vec}(\textrm{Conv}(\tW,\tX)) \nonumber\\
&= \begin{bmatrix}
     \textrm{vec}(\W_{(:,1,F_2)})^\textrm{T}\\
     \textrm{vec}(\W_{(:,2,F_2)})^\textrm{T}\\
     \dots\\
     \textrm{vec}(\W_{(:,O,F_2)})^\textrm{T}
     \end{bmatrix}
     \textrm{vec}(\X) \nonumber \\
&= \begin{bmatrix}
     \textrm{vec}(\U\V_{(:,1,F_2)})^\textrm{T}\\
     \textrm{vec}(\U\V_{(:,2,F_2)})^\textrm{T}\\
     \dots\\
     \textrm{vec}(\U\V_{(:,O,F_2)})^\textrm{T}
     \end{bmatrix}
     \textrm{vec}(\X) \nonumber \\
&= \begin{bmatrix}
     \textrm{vec}(\V_{(:,1,F_2)})^\textrm{T}\\
     \textrm{vec}(\V_{(:,2,F_2)})^\textrm{T}\\
     \dots \\
     \textrm{vec}(\V_{(:,O,F_2)})^\textrm{T}
     \end{bmatrix}
     \textrm{vec}(\U^\textrm{T}\X) \nonumber \\
&= \begin{bmatrix}
     \textrm{vec}(\V_{(:,1,F_2)})^\textrm{T}\\
     \textrm{vec}(\V_{(:,2,F_2)})^\textrm{T}\\
     \dots \\
     \textrm{vec}(\V_{(:,O,F_2)})^\textrm{T}
     \end{bmatrix}
     \textrm{vec}(\textrm{Conv}(\tU, \tX)) \nonumber \\
&= \textrm{vec}(\textrm{Conv}(\tV, \textrm{Conv}(\tU, \tX)) ).
\end{align}
Therefore,
$\textrm{Conv}(\tW,\tX)=\textrm{Conv}(\tV, \textrm{Conv}(\tU,\tX) )$.
\end{proof}

A more complicated case with different sizes of $\tX\in \bR^{H\times W \times I}$ and  strides $[s,s]$ can be proven in a similar way,
thus in the compressed CNN,
it is not necessary to reconstruct $\tW$ as in \cite{Garipov2016TTalike,wu2020hybrid} in the forward inference.
Instead, we only need to replace the original convolution layer with two consecutive factorized ones,
where the first has the weight tensor $\tU\in \bR^{F_1\times1\times I\times r}$ with strides $[s,1]$
and the second one $\tV \in \bR^{1\times F_2\times r\times O}$ with strides $[1,s]$.
In this manner,
CNNs are not only compressed but also accelerated in the inference period,
since the {\cb number of FLOPs in the convolutional layer drops from $H'W'F_1F_2IO$ to $H'WF_1Ir_m^r+H'W'F_2r_m^rO$,}
where $H'\times W'$ are the spatial sizes of the convolutional output.

After decomposition,
fine-tuning can be performed to recover performance.
{\cred
Although the fine-tuning of the compressed CNNs would consume some extra time,
it is acceptable since the process can be done quickly in high-performance servers with multiple GPUs.
In contrast, we pay more attention to the models' size, inference time, and accuracy.}
The whole process of RJSVD is shown in Algorithm \ref{RJSVD}.

The compression factor (\textbf{CF}) in this algorithm can be defined now.
Denoting the number of uncompressed parameters such as those in the BN layers and uncompressed layers as $\#\textrm{Other}$
where $\#$ means sizes of parameters,
the compression factor {\cbl can be calculated by}
\begin{align}\label{CF_rjsvd}
\textrm{CF}=\frac{\sum_{m=1}^{M}\sum_{n=1}^{N}\#\tW_m^n+\#\textrm{Other}}{\sum_{m=1}^{M}(\#\tV_m+\sum_{n=1}^{N}\#\tU_m^n)+\#\textrm{Other}}.
\end{align}

\begin{algorithm}
\caption{RJSVD}
\KwIn{A pre-trained CNN with weight tensors $\{\{\tW^n_m\}_{n=1}^N\}_{m=1}^{M}$, and target ranks $\{r_m^r\}_{m=1}^{M}$.}
\BlankLine
\For{$m=1,2,\cdots,M$}{
Obtain $\{\U_m^n\}_{n=1}^N, \V_m$ according to {\cbl formula} (\ref{rightshared}).\\
$\tV_m=\textrm{fold}(\V_m)$.\\
\For{$n=1,2,\cdots,N$}{
$\tU_m^n=\textrm{fold}(\U_m^n)$.\\
Replace the original convolutional layer $\tW^n_m$ with two compressed ones $\tU_m^n$ and $\tV_m$.
}
\label{RJSVD}
}

Fine-tune the compressed CNN.\\
\BlankLine
\KwOut{The compressed CNN and its weights $\tU_m^n, \tV_m$.}
\end{algorithm}

\subsubsection{Left Shared Joint SVD (LJSVD)}
In the RJSVD,
the right singular matrix is shared to further compress CNNs,
which enables weight typing across layers.
Following the same idea,
it is natural to derive \textbf{Left Shared Joint SVD} (\textbf{LJSVD}) by sharing the left factorized matrix,
which is:
\begin{align}
\textrm{Unfold}(\tW^n_m)=\W^n_m\approx\U_m\V^n_m,
\end{align}
where $\U_m\in  \bR^{F_1I \times r_m^l}$ are identical and shared for a group of $\tW^n_m, n=1,\cdots,N$,
while $\V^n_m\in \bR^{r_m^l\times F_2O}$ is different.
The decomposed structure derived by LJSVD is shown in the Figure \ref{fig:net}.

Similar to (\ref{optimization2}),
the optimization problem of LJSVD can be formulated as:
\begin{align}
\min_{\{\U^n_m\}^{N}_{n=1},\V_m}
\frac{1}{N}\sum_{n=1}^{N}||\W_{n,m} -\U_m\V^n_m||_2^2,  \label{optimization4}
\end{align}
which can be solved by performing truncated rank-$r_m^l$ SVD on the matrix produced by stacking $\W_m^n, n=1,2,\cdots, N$ horizontally:
\begin{align}
\U^m,
\begin{bmatrix}
    \V^1_m,
    \V^2_m,
    \cdots,
    \V^N_m
     \end{bmatrix}
=\textrm{SVD}_{r_m^l}(
\begin{bmatrix}
    \W^1_m,
    \W^2_m,
    \cdots,
    \W^N_m\\
     \end{bmatrix}).
     \label{leftshared}
\end{align}

The concrete steps are shown in the Algorithm \ref{ljsvd}.
With this algorithm,
the {\cb FLOPs} of a convolutional layer will drop from {\cb $H'W'F_1F_2IO$ to  $H'WF_1Ir_m^l+H'W'F_2r_m^lO$}.
And similar to (\ref{CF_rjsvd}),
the CF for LJSVD is
\begin{align}\label{CF_ljsvd}
\textrm{CF}=\frac{\sum_{m=1}^{M}\sum_{n=1}^{N}\#\tW_m^n+\#\textrm{Other}}{\sum_{m=1}^{M}(\#\tU_m+\sum_{n=1}^{N}\#\tV_m^n)+\#\textrm{Other}}.
\end{align}

\begin{algorithm}
\caption{LJSVD}
\KwIn{A pre-trained CNN with weight tensors $\{\{\tW^n_m\}_{n=1}^N\}_{m=1}^{M}$, and target ranks $\{r_m^l\}_{m=1}^{M}$.}
\BlankLine
\For{$m=1,2,\cdots,M$}{
Obtain $ \U_m, \{\V_m^n\}_{n=1}^N$ according to {\cbl formula} (\ref{leftshared}).\\
$\tU_m=\textrm{fold}(\V_m)$.\\
\For{$n=1,2,\cdots,N$}{
$\tV_m^n=\textrm{fold}(\V_m^n)$.\\
Replace the original convolutional layer $\tW^n_m$ with two compressed ones $\tU_m$ and  $\tV_m^n$.
}
}
Fine-tune the compressed CNN.\\
\BlankLine
\KwOut{The compressed CNN and its weights $\tU_m, \tV_m^n$.}
\label{ljsvd}
\end{algorithm}

\vline
\subsubsection{Binary Joint SVD (Bi-JSVD)}
The RJSVD and LJSVD jointly project the weight tensors of repeated layers into the same subspace by sharing the right factorized matrices or the left ones,
and in this section,
we combine RJSVD and LJSVD together to generate \textbf{Binary Joint SVD} (\textbf{Bi-JSVD}),
which is
\begin{align}
\textrm{Unfold}(\tW^n_m)=\W^n_m\approx \U^n_m\V_m+\U_m\V^n_m,
\label{bijsvd}
\end{align}
where $\U^n_m\in \bR^{F_1I\times r_m^r},\V_m\in \bR^{r_m^r\times F_2O},\U_m\in \bR^{F_1I\times r_m^l},\V^n_m\in \bR^{r_m^l\times F_2O}$.

{\cb
Note that Bi-JSVD is the generalization of RJSVD and LJSVD.
In other words,
RJSVD and LJSVD are particular cases of Bi-JSDV with $r_m^l=0$ and $r_m^r=0$, respectively.
For convenience,  we define the proportion of LJSVD in Bi-JSVD as
\begin{align}\label{p}
\bm{p}=\frac{r_m^l}{r_m^l+r_m^r}.
\end{align}
}

The optimization problem for (\ref{bijsvd}) can be formulated as:
\begin{align}
\min_{\{\U^n_m,\V^n_m\}^{N}_{n=1},\U_m,\V_m}
\frac{1}{N}\sum_{n=1}^{N}||\W_{n,m} -\U^n_m\V_m -\U_m\V^n_m||_2^2.  \label{optimization5}
\end{align}
{\cbl Since it is} inefficient to solve (\ref{optimization5}) with simultaneous optimization on  $\{\U^n_m\}^{N}_{n=1}, \V_m$, and $\U_m$,$\{\V^n_m\}^{N}_{n=1}$,
we solve it alternatively and iteratively as follows:
\BlankLine
\noindent(1) Given $\U_m,\{\V^n_m\}^{N}_{n=1}$,
let
\begin{align}
\widehat{\W_m^n}=\W_m^n - \U_m\V^n_m, \ \textrm{for} \ n=1,\cdots,N,
\label{res1}
\end{align}
then we have
\begin{align}
\begin{bmatrix}
    \U^1_m\\
    \U^2_m\\
    \cdots\\
    \U^N_m\\
     \end{bmatrix}
, \V_m
=\textrm{SVD}_{r_m^r}(
\begin{bmatrix}
    \widehat{\W^1_m}\\
    \widehat{\W^2_m}\\
    \cdots\\
    \widehat{\W^N_m}\\
     \end{bmatrix}).
     \label{Bi-rightshared}
\end{align}
\noindent(2) Given $\{\U^n_m\}^{N}_{n=1}, \V_m$,
let
\begin{align}
 \widehat{\widehat{\W_m^n}}=\W_m^n - \U^n_m \V_m, \ \textrm{for} \ n=1,\cdots,N,
 \label{res2}
\end{align}
then we have
\begin{align}
 \U_m,
\begin{bmatrix}
    \V^1_m,
    \V^2_m,
    \cdots,
    \V^N_m
     \end{bmatrix}
=\textrm{SVD}_{r_m^l}(
\begin{bmatrix}
    \widehat{\widehat{\W^1_m}},
     \widehat{\widehat{ \W^2_m}},
    \cdots,
     \widehat{\widehat{\W^N_m}}\\
     \end{bmatrix}
).\label{Bi-leftshared}
\end{align}
The decomposition will be converged by repeating the above steps $K$ times,
and then the fine-tuning of the compressed network can be used to recover its performance,
which is summarized in Algorithm \ref{Bi-JSVD}.
The CF for Bi-JSVD is
\begin{align}\label{CF_Bi-JSVD}
\small
\textrm{CF}=\frac{\sum_{m=1}^{M}\sum_{n=1}^{N}\#\tW_m^n+\#\textrm{Other}}{\sum_{m=1}^{M}(\#\tV_m+\#\tU_m+\sum_{n=1}^{N}(\#\tU_m^n+\#\tV_m^n))+\#\textrm{Other}},
\end{align}
and the complexity of a compressed convolutional layer is $\mathcal{O}((H'WF_1I +H'W'F_2O)(r_m^l+r_m^r))$.

\begin{algorithm}
\caption{Bi-JSVD}
\KwIn{A pre-trained CNN with weight tensors $\{\{\tW^n_m\}_{n=1}^N\}_{m=1}^{M}$, target ranks $\{r_m^r,r_m^l\}_{m=1}^{M}$, and iteration times $K$.}
\BlankLine

\For{$m=1,2,\cdots,M$}{
\BlankLine

Initialization: $\U_m=0,\quad\{\V_m^n\}_{n=1}^N=0$.\\
\For{$(k=0;k<K;k{++})$}{
Update $\{\U_m^n\}_{n=1}^N, \V_m$ according to (\ref{res1}) (\ref{Bi-rightshared}).\\
Update $\U_m, \{\V_m^n\}_{n=1}^N$ according to (\ref{res2}) (\ref{Bi-leftshared}).\\
}
\BlankLine
$\tU_m=\textrm{fold}(\U_m)$.\\
$\tV_m=\textrm{fold}(\V_m)$.\\
\For{$n=1,2,\cdots,N$}{
$\tU_m^n=\textrm{fold}(\U_m^n)$.\\
$\tV_m^n=\textrm{fold}(\V_m^n)$.\\
Replace the original convolutional layer $\tW^n_m$ with two parallel compressed sub-networks $\tU_m^n,\tV_m$ and  $\tU_m,\tV_m^n$ as shown in the Figure \ref{fig:net}.
}
}
Fine-tune the compressed CNN.\\
\BlankLine
\KwOut{The compressed CNN and its weights $\tU_m, \tV_m^n,\tU_m^n, \tV_m$.}
\label{Bi-JSVD}
\end{algorithm}

\newcommand\xrowht[2][0]{\addstackgap[.5\dimexpr#2\relax]{\vphantom{#1}}}
\newcommand{\tabincell}[2]{\begin{tabular}{@{}#1@{}}#2\end{tabular}}

\begin{table*}[]
	\centering
\begin{tabular}{c|c|c|c|c}
\hline
layer name       &output size         & ResNet-18                                                                          & ResNet-34        & ResNet-50                                                                  \\ \hline
 \tabincell{c}{conv1\\(original)}         &$32\times32$              & \multicolumn{3}{c}{$3\times 3$, 64, stride 1}                                                                                                                            \\ \hline
 \xrowht{26pt}
 \tabincell{c}{conv2\_x\\(original)}    &$32\times32$              & $\begin{bmatrix}3 \times 3,64 \\ 3 \times 3,64\end{bmatrix}\times 2$   & $\begin{bmatrix}3 \times 3,64 \\ 3 \times 3,64\end{bmatrix} \times 3$ & $\begin{bmatrix}1 \times 1,64 \\ 3 \times 3,64  \\1 \times 1,256  \end{bmatrix}\times3$
 \\ \hline
  \xrowht{26pt}
 \tabincell{c}{conv3\_x\\(decom)}     &$16\times16$             & $\begin{bmatrix}3 \times 3,128 \\ 3 \times 3,128\end{bmatrix} \times 2$ &$\begin{bmatrix}3 \times 3,128 \\ 3 \times 3,128\end{bmatrix} \times 4$ & $\begin{bmatrix}1 \times 1,128 \\ 3 \times 3,128  \\1 \times 1,512\end{bmatrix}\times4$
 \\ \hline
  \xrowht{26pt}
 \tabincell{c}{conv4\_x\\(decom)}     &$8\times8$             & $\begin{bmatrix}3 \times 3,256 \\ 3 \times 3,256\end{bmatrix} \times 2$ & $\begin{bmatrix}3 \times 3,256 \\ 3 \times 3,256\end{bmatrix} \times 6$  & $\begin{bmatrix}1 \times 1,256 \\ 3 \times 3,256 \\1 \times 1,1024\end{bmatrix}\times6$
 \\ \hline
  \xrowht{26pt}
 \tabincell{c}{conv5\_x\\(decom)}     &$4\times4$             & $\begin{bmatrix}3 \times 3,512 \\ 3 \times 3,512\end{bmatrix}\times 2$ & $\begin{bmatrix}3 \times 3,512 \\ 3 \times 3,512\end{bmatrix} \times 3$ & $\begin{bmatrix}1 \times 1,512 \\ 3 \times 3,512  \\1 \times 1,2048\end{bmatrix}\times3$
 \\ \hline
\multicolumn{5}{c}{average pool, 10-d fc for CIFAR-10/100-d fc for CIFAR-100, softmax}                                                                                                                                                \\ \hline
\end{tabular}
\caption{The architectures of CNNs for CIFAR-10 and CIFAR-100. Layers remarked with ``original'' will not be decomposed in the following experiments, while those remarked with ``decom'' will be decomposed. }
\label{cnns}
\end{table*}

\section{Experiments}
\label{sec:experiments}
To validate the effectiveness of the proposed algorithms, we conduct extensive experiments on various benchmark data sets and several widely used networks with different depths. The proposed algorithms are adopted to compress the networks, and we compare the results with some state-of-the-art decomposition-based compression methods. All experiments are performed on one TITAN Xp GPU under Tensorflow1.15 and Ubuntu18.04.$\footnote{The codes are available at \url{https://github.com/ShaowuChen/JointSVD}.}$

\subsection{Evaluation on CIFAR-10 and CIFAR-100}
\label{Evaluation on CIFAR-10 and CIFAR-100}

\subsubsection{Overall settings}
\label{Overall settings}
\paragraph{Datasets}
In this section, we evaluate the proposed methods on CIFAR-10 and CIFAR-100 \cite{Cifar-10Dataset}. Both data sets have 60,000 $32\times32\times3$ images, including 50,000 training images and 10,000 testing images. The former contains 10 classes, while the latter includes 100 categories, and thus is more challenging for classification. For data preprocessing, all the images are normalized with $mean=[0.4914, 0.4822, 0.4465]$ and standard deviation $std=[0.2023,0.1994,0.2010]$. For data augmentation, a $32\times32$ random crop is adopted on the zero-padded $40\times40$ training images followed by a random horizontal flip.

\paragraph{Networks}
\label{Networks}
We evaluate the proposed method on the widely used ResNet with various depths \textit{i.e.}, ResNet-18, ResNet-34, ResNet-50. There are two main reasons for choosing them: (1) the widely used ResNet is a typical representative of architectures that adopt the repeated module design; (2) by compressing these compact networks, the ability of the proposed algorithms can be clearly presented \cite{PPCA}. The architectures and Top-1 accuracy of the baseline CNNs are shown in Tables \ref{cnns} and \ref{CNNsAcc}, respectively. We randomly initialize the weights using the truncated normal initializer by setting the standard deviation to 0.01. The SGD optimizer with a momentum of 0.9 and a weight decay of 5e-4 is used to train the CNNs for 300 epochs. The batch size for ResNet-18 and ResNet-34 is 256, while it is 128 for ResNet-50 due to the limitation of graphics memory. The learning rate starts from 0.1 and is divided by 10 in the 140th, 200th, and 250th epochs.

\begin{table}[htbp]
\normalsize
    \sisetup{table-format=2.2} 
    \centering
    \begin{tabular}{@{} c S S S S @{}}
    \toprule
     & \multicolumn{2}{c@{}}{CIFAR-10} & \multicolumn{2}{c@{}}{CIFAR-100}\\
    \cmidrule(lr){2-3} \cmidrule(l){4-5}
    CNN& {Acc. (\%)} & {Size (M)} & {Acc. (\%)} & {Size (M)}\\
    \midrule
    ResNet-18 & 94.80 &11.16    & 70.73 &11.21     \\
    ResNet-34 & 95.11 &21.27    & 75.81 &21.31     \\
    ResNet-50 & 95.02 &23.50    & 75.67 &23.69     \\
    \bottomrule
    \end{tabular}
    \caption{Top-1 accuracies and parameter sizes of the baseline CNNs on CIFAR-10 and CIFAR-100. ``Acc.'' means ``Accuracy''.}
    \label{CNNsAcc}%
\end{table}

\paragraph{Methods for comparison}
To evaluate the effectiveness of the proposed algorithms, three state-of-the-art methods, including one matrix-decomposition-based method, Tai \textit{et al.}\cite{Tai2016Convolutional}, and two tensor-decomposition-based ones, Tucker \cite{Kim2016Tucker} and NC\_CTD \cite{JSTSP}, are employed for comparison.

\paragraph{CF and FLOPs}
When setting CFs, we take \cite{Tai2016Convolutional} as the anchor. An equal proportion is set for each layer to determine the ranks for \cite{Tai2016Convolutional} and obtain the CF, with which the ranks for Tucker \cite{Kim2016Tucker}, NC\_CTD \cite{JSTSP} and our methods are then calculated under the same CFs. In NC\_CTD \cite{JSTSP}, since the proportion of the shared component and independent component affects the final performance, without loss of generality, we set it to $1:1$ and $2:1$ and report the better result only. We use the ``TensorFlow.profiler'' function to calculate the FLOPs which indicates the theoretical acceleration of the compressed models.

\subsubsection{Tuning the Parameter $K$}
\label{sec:tuning}
In the beginning, we first determine the iteration number $K$ for Bi-JSVD since it might affect the quality of the initialization after decomposition and before fine-tuning. Empirically, a more accurate approximation in decomposition would bring better performance after fine-tuning, making $K$ an inconspicuous but vital hyper-parameter. A large $K$, {\cb such as 100, can ensure the convergence of decomposition, but it would consume more time.} To find a suitable $K$, we conduct experiments on ResNet-34 for CIFAR-10, and {\cred take} the raw accuracy of the decomposed networks before fine-tuning as the {\cred evaluation criterion}.

\paragraph{Settings} We decompose the sub-networks {\cbl ``}conv$i$\_x{\cbl" } for $i=3,4,5$ in Table \ref{cnns} except {\cbl for} the conv1 and conv2\_x {\cbl since they have much fewer} parameters {\cbl comparing with the other sub-networks}, {\cbl and} decomposing them would bring relatively severe accumulated error. In each sub-networks ``conv$i$\_x'' for $i=3,4,5$, the first convolutional layer of the first block has only half the input depth (\textbf{HID layer}) compared with the one in other blocks, thus they are not {\cbl decomposed by} Bi-JSVD {\cbl jointly} but the matrix method \cite{Tai2016Convolutional} separately. The CF and iteration times are set to 13.9 and $K\in\{10,30,50,70\}$, respectively, and we set $\bm{p}$, the proportion of LJSVD, to $p= 0.3, 0.5, 0.7,0.9$.

\paragraph{Results and analysis}
The results are shown in Table \ref{tuningK}. It is shown that $K=30$ and $K=70$ would bring relatively higher raw accuracy, and the gaps between them are insignificant. Therefore, $K$ will be set to 30 in {\cb all} the following experiments to obtain high raw accuracy with less complexity. Furthermore, it seems that the proportion of LJSVD or RJSVD in Bi-JSVD indicated by $p$  would affect the raw accuracy of the compressed network. To give a conclusion on how $p$ affects the final performance, more experiments are needed, and we will discover it in the following sections.

\begin{table}[htbp]
\normalsize
    \centering
    \begin{tabular}{ccccc}
    \toprule
     & \multicolumn{4}{c}{$p$}\\
     \cmidrule{2-5}
     $K$ & 0.3 & 0.5 & 0.7 & 0.9 \\
    \midrule
    10 & 57.64\% & 53.44\%   & 51.04\% & 42.69\%  \\
    30 & \textcolor{red}{\textbf{58.12\%}} & 56.43\%   & \textcolor{red}{\textbf{51.75\%}} & 44.01\%  \\
    50 & 57.33\% & 56.77\%   & 50.23\% & 43.45\%  \\
    70 & 57.65\% & \textcolor{red}{\textbf{56.85\%}}   & 49.47\% & \textcolor{red}{\textbf{44.13\%}}  \\
    \bottomrule
    \end{tabular}
\caption{Raw accuracies of compressed ResNet-34 with Bi-JSVD on CIFAR-10 before fine-tuning.}
\label{tuningK}
\end{table}

\begin{table}[h]
\setlength\tabcolsep{2pt}
  \centering
    \begin{tabular}{cclcccc}
    \toprule
    \tabincell{c}{CNN\\\& acc.\\{\cb{\& FLOPs}}} & \tabincell{c}{CF\\($\times$)} & Method & \tabincell{c}{Raw\\acc. (\%)} &\tabincell{c}{Acc.  (\%)\\(mean$\pm$std)} & \tabincell{c}{Best\\acc. (\%)}  & \tabincell{c}{\cb{FLOPs}}\\
\midrule
    \multirow{18}[4]{*}{ \tabincell{c}{ResNet-18\\\\94.80\%\\\\{\cb{11.11E8}}}} & \multirow{9}[2]{*}{17.76}
    &       Tai \textit{et al.}\cite{Tai2016Convolutional} & 10.31 & 92.49$\pm$0.23 & 92.93 & {\cb{3.42E8}} \\
    &     & Tucker \cite{Kim2016Tucker} & \textcolor{red}{\textbf{23.77}} & 91.93$\pm$0.13 & 92.41 &{\cb{3.41E8}} \\
    &     &NC\_CTD \cite{JSTSP}    &23.51  & 92.20$\pm$0.29  &92.80 &{\cb{3.47E8}}\\
    \cdashline{3-7}
    \noalign{\vskip 0.5ex}
        &     & LJSVD   & 11.44 & 92.88$\pm$0.08 & 93.00 &{\cb{3.45E8}}  \\
        &     & RJSVD-1 & 10.00 & \textcolor{red}{\textbf{93.19$\pm$0.04}} & \textcolor{red}{\textbf{93.36}} &{\cb{3.50E8}} \\
        &     & RJSVD-2& 10.27 & 92.75$\pm$0.09 & 93.09 &{\cb{3.45E8}} \\
        &     & Bi-JSVD0.3 & 13.76 & 92.81$\pm$0.06 & 92.91 &{\cb{3.45E8}}\\
        &     & Bi-JSVD0.5 & 13.09 & 92.70$\pm$0.11 & 92.80 &{\cb{3.44E8}} \\
        &     & Bi-JSVD0.7 & 13.64 & 93.11$\pm$0.06 & 93.32 &{\cb{3.45E8}}\\
\cmidrule{2-7}
        & \multirow{8}[2]{*}{11.99} & Tai \textit{et al.}\cite{Tai2016Convolutional} & 54.42 & 93.55$\pm$0.08 & 93.83 & {\cb{3.65E8}}\\
        &     & Tucker \cite{Kim2016Tucker} & \textcolor{red}{\textbf{57.51}} & 92.42$\pm$0.29 & 92.84 & {\cb{3.63E8}}\\
        &     &NC\_CTD \cite{JSTSP}    & 54.15   & 92.85$\pm$0.25  &93.34 & {\cb{3.74E8}}\\
    \cdashline{3-7}
    \noalign{\vskip 0.5ex}
        &     & LJSVD& 50.15 & 93.62$\pm$0.10 & 93.98  & {\cb{3.73E8}}\\
        &     & RJSVD-1 & 30.53 & 93.75$\pm$0.07 &  \textcolor{red}{\textbf{94.22}} & {\cb{3.83E8}}\\
        &     & RJSVD-2& 42.43 & 93.47$\pm$0.12 & 93.80 & {\cb{3.73E8}} \\
        &     & Bi-JSVD0.3 & 52.84 & 93.77$\pm$0.17 & 94.15 & {\cb{3.73E8}} \\
        &     & Bi-JSVD0.5 & 48.45 & 93.49$\pm$0.07 & 93.73 & {\cb{3.72E8}}\\
        &     & Bi-JSVD0.7 & 52.31 & \textcolor{red}{\textbf{93.84$\pm$0.09}} & 94.16  & {\cb{3.73E8}}\\

\midrule
    \multirow{18}[3]{*}{\tabincell{c}{ResNet-34\\\\95.11\%\\\\{\cb{23.19E8}}}} & \multirow{8}[2]{*}{22.07} & Tai \textit{et al.}\cite{Tai2016Convolutional} & 12.93 & 93.66$\pm$0.13 & 93.98 & {\cb{5.20E8}}\\
    &     & Tucker \cite{Kim2016Tucker} & \textcolor{red}{\textbf{19.73}} & 92.83$\pm$0.07 & 93.06 & {\cb{5.20E8}}\\
    &     &NC\_CTD \cite{JSTSP}    &19.12    & 92.98$\pm$0.15  &93.35 & {\cb{5.71E8}} \\
    \cdashline{3-7}
    \noalign{\vskip 0.5ex}
        &     & LJSVD& 15.96 & \textcolor{red}{\textbf{93.97$\pm$0.10}} & 94.07 & {\cb{5.47E8}} \\
        &     & RJSVD-1 & 10.21 & 93.82$\pm$0.07 & 94.00 & {\cb{5.54E8}}\\
        &     & RJSVD-2& 12.97 & 93.77$\pm$0.03 & 93.95  & {\cb{5.47E8}}\\
        &     & Bi-JSVD0.3 & 17.38 & 93.66$\pm$0.05 & 93.88 & {\cb{5.44E8}}\\
        &     & Bi-JSVD0.5 & 16.87 & 93.69$\pm$0.09 & 93.98 & {\cb{5.43E8}}\\
        &     & Bi-JSVD0.7 & 15.83 & 93.77$\pm$0.13 & \textcolor{red}{\textbf{94.09}} & {\cb{5.44E8}}\\
\cmidrule{2-7}        & \multirow{8}[1]{*}{13.92}
               & Tai \textit{et al.}\cite{Tai2016Convolutional} & 49.34 & 93.95$\pm$0.05 & 94.12 & {\cb{5.71E8}}\\
        &     & Tucker \cite{Kim2016Tucker} & \textcolor{red}{\textbf{64.13}} & 93.04$\pm$0.21 & 93.39  & {\cb{5.70E8}}\\
        &     &NC\_CTD \cite{JSTSP}    & 60.12 & 93.30$\pm$0.13 &93.69 & {\cb{6.16E8}}\\
    \cdashline{3-7}
    \noalign{\vskip 0.5ex}
        &     & LJSVD& 39.53 & \textcolor{red}{\textbf{94.39$\pm$0.15}} & 94.61 & {\cb{6.27E8}} \\
        &     & RJSVD-1 & 46.38 & 94.23$\pm$0.23 & \textcolor{red}{\textbf{94.73}} & {\cb{6.42E8}}  \\
        &     & RJSVD-2& 45.60 & 93.94$\pm$0.06 & 94.12 & {\cb{6.27E8}} \\
        &     & Bi-JSVD0.3 & 58.12 & 94.17$\pm$0.09 & 94.41 & {\cb{6.22E8}}\\
        &     & Bi-JSVD0.5 & 56.43 & 94.08$\pm$0.06 & 94.32  & {\cb{6.24E8}}\\
        &     & Bi-JSVD0.7 & 51.75 & 94.02$\pm$0.19 & 94.34 & {\cb{6.22E8}}\\
\midrule
    \multirow{14}[3]{*}{\tabincell{c}{ResNet-50\\\\95.02\%\\\\ {\cb{25.96E8}}  }} & \multirow{7}[1]{*}{6.48} & Tai \textit{et al.}\cite{Tai2016Convolutional} & 20.24 & 92.38$\pm$0.31 & 92.96  & {\cb{7.19E8}}\\
    \cdashline{3-7}
    \noalign{\vskip 0.5ex}
        &     & LJSVD& 12.07 & 93.08$\pm$0.17 & 93.37 & {\cb{7.63E8}}\\
        &     & RJSVD-1 & 13.85 & \textcolor{red}{\textbf{93.28$\pm$0.09}} & \textcolor{red}{\textbf{93.58}}  & {\cb{7.77E8}} \\
        &     & RJSVD-2& \textcolor{red}{\textbf{20.48}} & 92.75$\pm$0.28 & 93.26  &{\cb{7.73E8}}\\
        &     & Bi-JSVD0.3 & 16.11 & 92.97$\pm$0.12 & 93.30   & {\cb{7.66E8}}\\
        &     & Bi-JSVD0.5 & 13.46 & 92.90$\pm$0.15 & 93.19  & {\cb{7.66E8}}\\
        &     & Bi-JSVD0.7 & 10.73 & 92.84$\pm$0.26 & 93.49  & {\cb{7.61E8}}\\
\cmidrule{2-7}        & \multirow{7}[2]{*}{5.37}
       & Tai \textit{et al.}\cite{Tai2016Convolutional} & 28.40 & 92.99$\pm$0.42 & 93.44  & {\cb{7.97E8}}\\
    \cdashline{3-7}
    \noalign{\vskip 0.5ex}
        &     & LJSVD& 22.67 & \textcolor{red}{\textbf{93.39$\pm$0.18}} & \textcolor{red}{\textbf{93.81}}   & {\cb{8.87E8}}\\
        &     & RJSVD-1 & 24.94 & 92.97$\pm$0.10 & 93.33  & {\cb{9.17E8}} \\
        &     & RJSVD-2&34.46 & 93.24$\pm$0.11 & 93.41   & {\cb{9.11E8}}\\
        &     & Bi-JSVD0.3 & 32.18 & 92.86$\pm$0.19 & 93.17   & {\cb{8.99E8}}\\
        &     & Bi-JSVD0.5 & 32.39 & 93.08$\pm$0.21 & 93.50  & {\cb{8.95E8}}\\
        &     & Bi-JSVD0.7 &  \textcolor{red}{\textbf{35.34}} & 93.06$\pm$0.09 & 93.28  & {\cb{8.89E8}}\\
    \bottomrule
    \end{tabular}%
   \caption{Comparison of compressed ResNet following the ``pre-train$\to$decompose$\to$fine-tune'' pipeline on CIFAR-10.
   ``Raw acc.'' means accuracy before fine-tuning.
   ``Acc.'' and ``Best acc.'' represent the average accuracy and best accuracy after fine-tuning among repeated experiments, respectively.}
   \label{results11}%
\end{table}%

\begin{table}[h]
\setlength\tabcolsep{2pt}
  \centering
    \begin{tabular}{cclcccc}
    \toprule
    \tabincell{c}{CNN\\\& acc.\\{\cb{\& FLOPs}}} & \tabincell{c}{CF\\($\times$)} & Method & \tabincell{c}{Raw\\acc. (\%)} &\tabincell{c}{Acc.  (\%)\\(mean$\pm$std)} &\tabincell{c}{Best \\acc. (\%)} & \tabincell{c}{\cb{FLOPs}}\\
    \midrule
     \multirow{18}[4]{*}{\tabincell{c}{ResNet-18\\\\70.73\%\\\\{\cb{11.11E8}}}} & \multirow{8}[2]{*}{16.61} & Tai \textit{et al.}\cite{Tai2016Convolutional} & 2.07 & 71.15$\pm$0.16 & 71.86 & {\cb{3.42E8}}\\
   &     & Tucker \cite{Kim2016Tucker} & \textcolor{red}{\textbf{7.69}} & 71.54$\pm$0.26 & 72.62 & {\cb{3.41E8}}\\
   &    & NC\_CTD \cite{JSTSP} &3.94   &\textcolor{red}{\textbf{72.02$\pm$0.13}}&72.74 & {\cb{3.47E8}}\\
    \cdashline{3-7}
    \noalign{\vskip 0.5ex}
       &     & LJSVD& 2.43 & 71.63$\pm$0.29 & 72.36 & {\cb{3.45E8}}\\
       &     & RJSVD-1 & 1.67 & \textcolor{red}{\textbf{72.02$\pm$0.30}} & \textcolor{red}{\textbf{72.87}} & {\cb{3.50E8}}\\
       &     & RJSVD-2& 2.97 & 71.35$\pm$0.19 & 71.78 & {\cb{3.45E8}}\\
      &      &Bi-JSVD0.3 & 3.78 & 71.56$\pm$0.13 & 72.04 & {\cb{3.45E8}}\\
      &     & Bi-JSVD0.5 & 4.14 & 71.93$\pm$0.21 & 72.55 & {\cb{3.44E8}}\\
        &     & Bi-JSVD0.7 & 3.95 & 71.61$\pm$0.11 & 72.28 & {\cb{3.45E8}}\\

     \cmidrule{2-7}& \multirow{8}[2]{*}{11.47} & Tai \textit{et al.}\cite{Tai2016Convolutional} & 9.22 & 73.63$\pm$0.17 & 74.42 & {\cb{3.65E8}}\\
       &     & Tucker \cite{Kim2016Tucker} & \textcolor{red}{\textbf{26.33}} & 72.24$\pm$0.17 & 72.78 & {\cb{3.64E8}}\\
       &    & NC\_CTD \cite{JSTSP} &22.02   &72.88$\pm$0.11 &73.46 & {\cb{3.68E8}}\\
    \cdashline{3-7}
    \noalign{\vskip 0.5ex}
       &     & LJSVD   & 6.42 & \textcolor{red}{\textbf{74.16$\pm$0.33}} & \textcolor{red}{\textbf{75.02}} & {\cb{3.73E8}}\\
       &     & RJSVD-1 & 5.95 & 74.00$\pm$0.10 & 74.27 & {\cb{3.83E8}}\\
       &     & RJSVD-2& 8.88 & 73.71$\pm$0.14 & 74.17 & {\cb{3.73E8}}\\
       &     & Bi-JSVD0.3 & 7.79 & 74.00$\pm$0.13 & 74.43 &{\cb{3.73E8}}\\
       &     & Bi-JSVD0.5 & 8.31 & 73.71$\pm$0.17 & 73.93 & {\cb{3.72E8}}\\
       &     & Bi-JSVD0.7 & 7.77 & 73.76$\pm$0.15 & 74.21 & {\cb{3.73E8}}\\

    \midrule
    \multirow{18}[3]{*}{\tabincell{c}{ResNet-34\\\\75.81\%\\\\{\cb{23.19E8}}}} & \multirow{8}[2]{*}{21.11} & Tai \textit{et al.}\cite{Tai2016Convolutional} & 2.30 & 73.30$\pm$0.15 & 73.56 & {\cb{5.20E8}}\\
       &     & Tucker \cite{Kim2016Tucker} & \textcolor{red}{\textbf{5.66}} & 73.53$\pm$0.14 & 73.81 & {\cb{5.20E8}}\\
       &    & NC\_CTD \cite{JSTSP} &3.77   &73.91$\pm$0.27 &74.59 &{\cb{5.71E8}}\\
    \cdashline{3-7}
    \noalign{\vskip 0.5ex}
       &     & LJSVD& 4.75 & \textcolor{red}{\textbf{74.39$\pm$0.02}} & \textcolor{red}{\textbf{74.92}} & {\cb{5.47E8}}\\
       &     & RJSVD-1 & 3.45 & 73.96$\pm$0.16 & 74.31 & {\cb{5.54E8}}\\
       &     & RJSVD-2& 4.33 & 73.81$\pm$0.13 & 74.33 & {\cb{5.47E8}}\\
      &     & Bi-JSVD0.3 & 3.29 & 73.88$\pm$0.15 & 74.38 & {\cb{5.44E8}}\\
      &     & Bi-JSVD0.5 & 5.12 & 74.03$\pm$0.12 & 74.55 & {\cb{5.43E8}}\\
       &     & Bi-JSVD0.7 & 5.19 & 74.05$\pm$0.31 & 74.62 & {\cb{5.44E8}}\\

\cmidrule{2-7}& \multirow{8}[1]{*}{13.55} & Tai \textit{et al.}\cite{Tai2016Convolutional} & 6.35 & 75.29$\pm$0.23 & 75.50 & {\cb{5.71E8}}\\
       &     & Tucker \cite{Kim2016Tucker} & \textcolor{red}{\textbf{19.69}} & 74.07$\pm$0.44 & 75.18 & {\cb{5.70E8}}\\
       &    & NC\_CTD \cite{JSTSP} &17.51  &74.53$\pm$0.12   &75.18 & {\cb{6.67E8}}\\
    \cdashline{3-7}
    \noalign{\vskip 0.5ex}
       &     & LJSVD& 10.66 & \textcolor{red}{\textbf{75.84$\pm$0.16}} & 76.26 & {\cb{6.27E8}}\\
        &     & RJSVD-1 & 6.09 & 75.43$\pm$0.07 & 75.76 & {\cb{6.42E8}}\\
     &     & RJSVD-2& 10.83 & 75.34$\pm$0.37 & 75.90 & {\cb{6.27E8}}\\
        &     & Bi-JSVD0.3 & 13.52 & \textcolor{red}{\textbf{75.84$\pm$0.21}} & \textcolor{red}{\textbf{76.48}} & {\cb{6.22E8}}\\
        &     & Bi-JSVD0.5 & 13.47 & 75.61$\pm$0.11 & 75.85 & {\cb{6.25E8}}\\
       &     & Bi-JSVD0.7 & 10.73 & \textcolor{red}{\textbf{75.84$\pm$0.37}} & 76.47 & {\cb{6.22E8}}\\

\midrule
    \multirow{14}[3]{*}{\tabincell{c}{ResNet-50\\\\75.67\%\\\\{\cb{25.96E8}}}} & \multirow{7}[1]{*}{6.21} & Tai \textit{et al.}\cite{Tai2016Convolutional} & 7.65 & 74.36$\pm$0.23 & 74.90 & {\cb{7.20E8}}\\
    \cdashline{3-7}
    \noalign{\vskip 0.5ex}
        &     & LJSVD& 3.71 & 75.04$\pm$0.32 & 75.63 & {\cb{7.63E8}}\\
        &     & RJSVD-1 & 2.18 & 73.56$\pm$0.41 & 74.52 & {\cb{7.78E8}}\\
        &     & RJSVD-2& 2.66 & 73.51$\pm$0.55 & 74.59 & {\cb{7.74E8}}\\
      &     & Bi-JSVD0.3 & 3.00   & 74.79$\pm$0.14 & 75.14 & {\cb{7.66E8}}\\
       &     & Bi-JSVD0.5 & 2.65 & 74.25$\pm$0.27 & 74.65 & {\cb{7.66E8}}\\
        &     & Bi-JSVD0.7 & 3.34 & 74.47$\pm$0.09 & 74.92 & {\cb{7.61E8}}\\

\cmidrule{2-7}&\multirow{7}[2]{*}{5.19} & Tai \textit{et al.}\cite{Tai2016Convolutional} & \textcolor{red}{\textbf{24.67}} & 75.09$\pm$0.31 & 75.71 & {\cb{7.97E8}}\\
    \cdashline{3-7}
    \noalign{\vskip 0.5ex}
        &     & LJSVD& 5.57 & \textcolor{red}{\textbf{75.67$\pm$0.45}} & \textcolor{red}{\textbf{76.64}} & {\cb{8.88E8}}\\
        &     & RJSVD-1 & 4.72 & 74.43$\pm$0.46 & 75.49 & {\cb{9.17E8}}\\
        &     & RJSVD-2& 8.02 & 74.36$\pm$0.33 & 74.81 & {\cb{9.11E8}}\\
        &     & Bi-JSVD0.3 & 6.71 & 75.30$\pm$0.41 & 76.06 & {\cb{9.00E8}}\\
        &     & Bi-JSVD0.5 & 5.76 & 74.81$\pm$0.62 & 75.81 & {\cb{8.95E8}}\\
        &     & Bi-JSVD0.7 & 5.06 & 75.01$\pm$0.30 & 75.93 & {\cb{8.90E8}}\\
    \bottomrule
    \end{tabular}%
   \caption{Comparison of compressed ResNet following the ``pre-train$\to$decompose$\to$fine-tune'' pipeline on CIFAR-100.}
   \label{results12}%
\end{table}

\subsubsection{Performance Comparison}
\label{PC}

\paragraph{Settings} We do not decompose conv1 and conv2\_x in all CNNs  as it is discussed in \ref{sec:tuning}. In ResNet-18 and ResNet-34, we compare the proposed methods with Tai \textit{et al.}\cite{Tai2016Convolutional}, Tucker \cite{Kim2016Tucker}, and NC\_CTD \cite{JSTSP} following the ``pre-train$\to$decompose$\to$fine-tune'' pipeline.
{\cred
Whereas, in ResNet-50, there are lots of $1\times1$ convolutional layers with weight tensors of dimensions $1\times1\times I\times O$ that make the Tucker or tensor decomposition equal to matrix decomposition, thus only the Tai \textit{et al.} \cite{Tai2016Convolutional} is used for comparing.} To see the gap of performance clearly, the CF in ResNet-18 and ResNet-34 is set to be relatively large and roughly from 14 to 20. {\cred However, we set the CF for ResNet-50 as 5--6 since ResNet-50 with BottleNecks is very compact.}

{\cred
The HID layers with half the input depth are also decomposed for all the methods. For consistency, as in the Section \ref{sec:tuning}, the HID layers in LJSVD and Bi-JSVD will be decomposed via \cite{Tai2016Convolutional} separately, since LJSVD and Bi-JSVD are not compatible with the HID layers. However, RJSVD stacks folded weights vertically, thus the HID layers can be included for decomposing jointly. To evaluate RJSVD comprehensively and fairly,
we implement \textbf{RJSVD-1} and \textbf{RJSVD-2} that decompose the HID layers jointly and separately, respectively.}
To answer the query posted in the last experiment, \textit{i.e.}, how $p$ in Bi-JSVD affects the performance of the compress networks, we set different $p$ and name the methods as \textbf{Bi-JSVD$\bm{p}$}, for $p=0.3,0.5,0.7$. {\cb Note that here $\textrm{RJSVD-2}$, $\textrm{LJSVD}$ are equivalent to $\textrm{Bi-JSVD0}$ and $\textrm{Bi-JSVD1}$, respectively.}

To conduct a fair comparison and avoid early convergence, we use the same training settings in the fine-tuning stage as the baseline CNNs in Section \ref{Overall settings} for all the methods {\cred (including the methods for comparison) without particular tuning and tricks}, except that the batch size for all networks is set to 128 {\cred for uniformity}.
There are 12 groups of experiments implemented in total, and all the experiments are {\cbl repeated} three times to obtain average performance. {\cred We also present the highest accuracy recorded during fine-tuning, since they indicate the potential of the corresponding models.}

\begin{figure}
    \centering
    \includegraphics[width=1\linewidth]{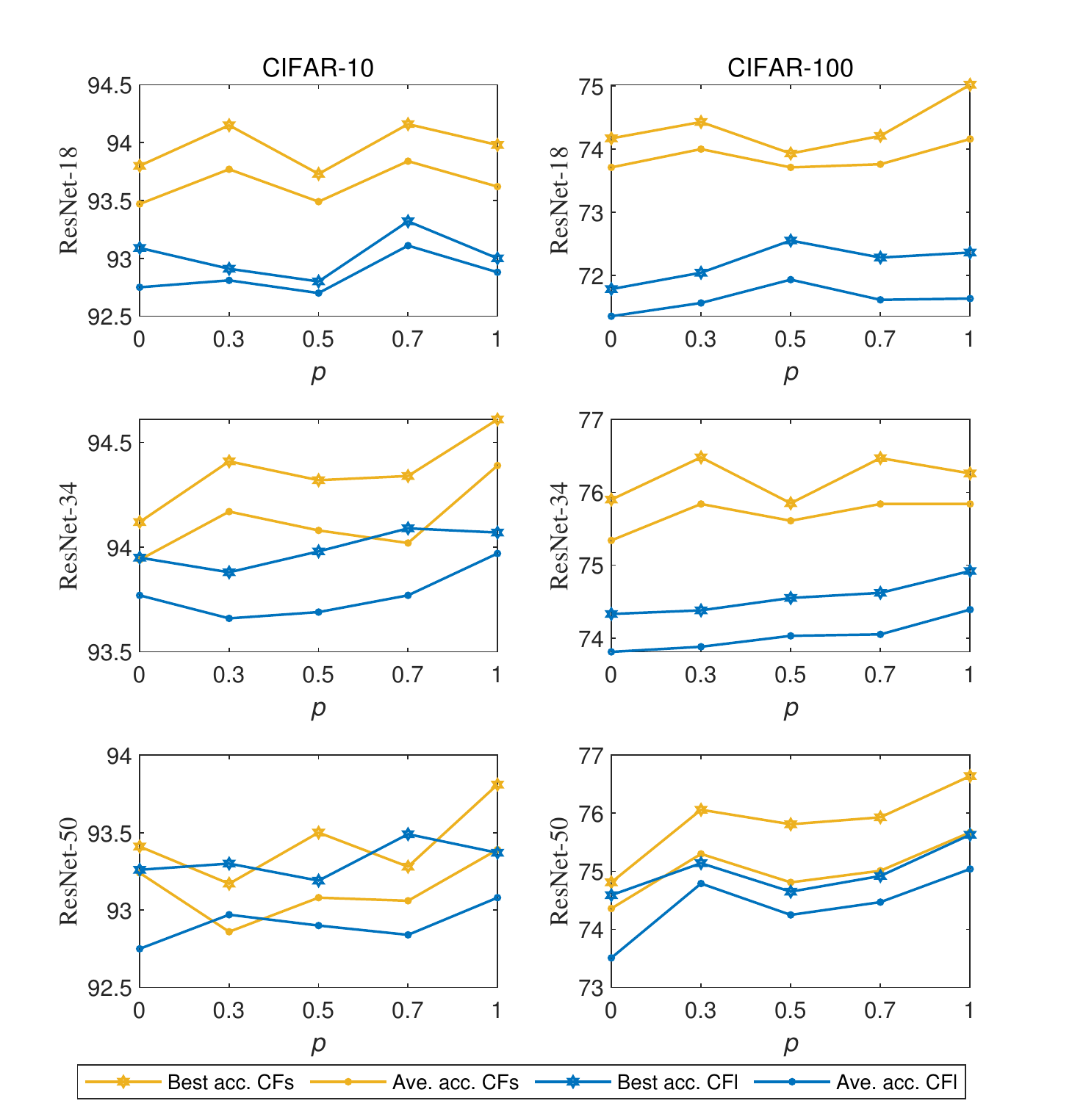}
    \caption{The performance of Bi-JSVD for pre-trained CNNs under different $p$. ``Ave. acc.'',  ``Best acc.'', ``CFs'' and ``CFl'' {\cbl mean} Average accuracy, Best accuracy, the smaller CF and the larger one for each CNN shown in the TABLE \ref{results11} and \ref{results12}, respectively.}
    \label{line1}
\end{figure}

\begin{figure}[h]
    \centering
    \includegraphics[width=1\linewidth]{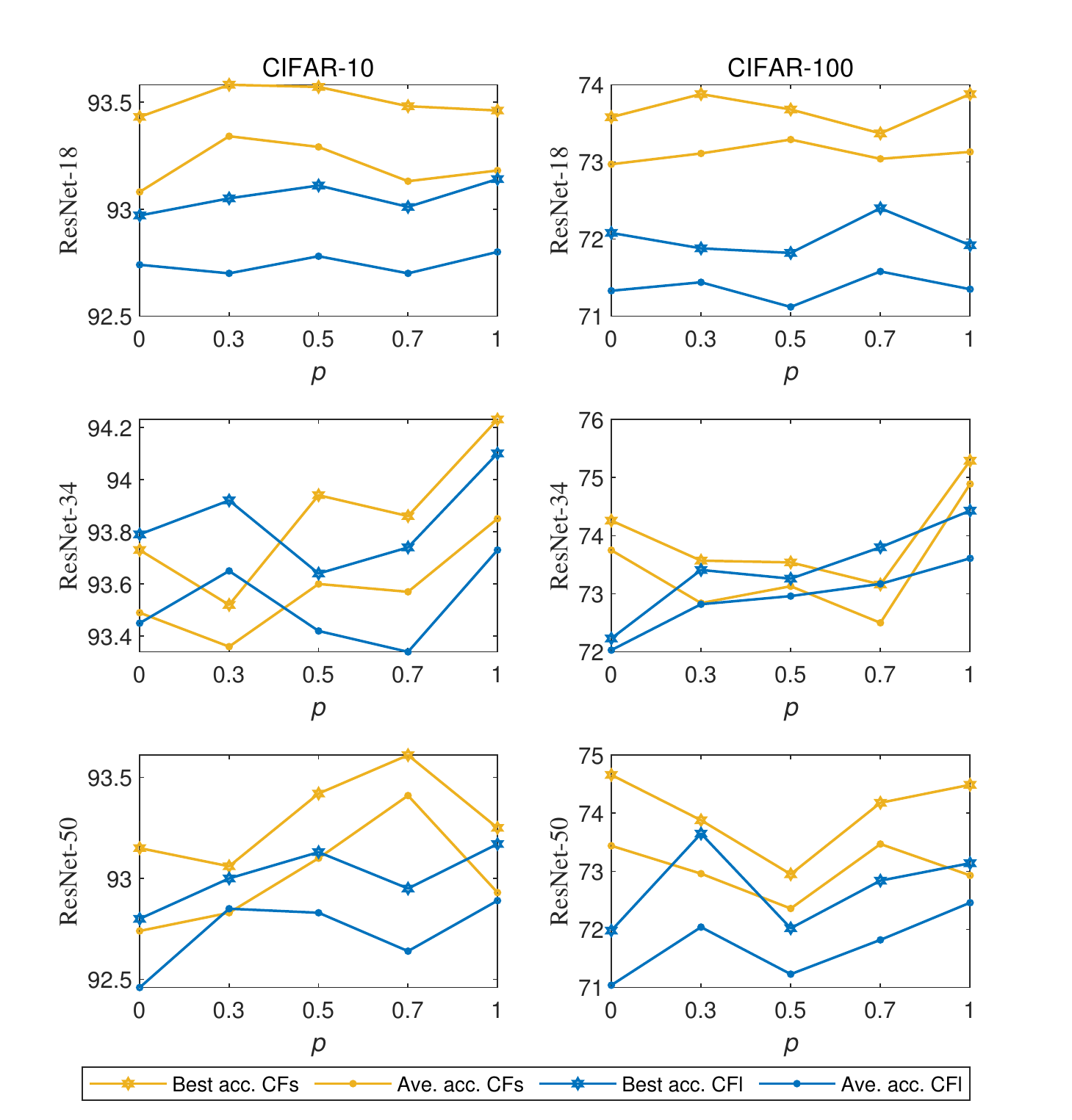}
    \caption{The performance of Bi-JSVD for CNNs trained from scratch under different $p$. ``Ave. acc.'',  ``Best acc.'', ``CFs'' and ``CFl'' {\cbl mean} Average accuracy, Best accuracy, the smaller CF and the larger one for each CNN shown in the TABLE \ref{results21} and \ref{results22}, respectively.}
    \label{line2}
\end{figure}

\paragraph{Results}
The results are shown in Tables \ref{results11} and \ref{results12}. To observe how $p$ affects Bi-JSVD, we plot the curves of the average and best performance with $p=[0,0.3,0.5,0.7,1]$ under different CF in Figures \ref{line1} and \ref{line2}.

\paragraph{Analysis}
From the results, we summarize {and analyze} several important observations:
\begin{enumerate}[(i)]
  \item\label{obsevation1}
  All the proposed methods outperform the compared methods after fine-tuning in most cases. CFs such as $22\times$ much higher than previous works are achieved with relatively slight performance degradation, demonstrating the effectiveness of the proposed joint methods. Moreover, it is shown that a deeper network with more layers for joint decomposition would expand the advantages of the proposed methods in alleviating performance degradation. Besides, the advantages of our methods are further expanded in the more challenging CIFAR-100 tasks. For example, the accuracy of the ResNet-18 jointly decomposed by LJSVD on CIFAR-100 under $\textrm{CF}=11.47$   is 1.92\% higher than Tucker and 0.53\% higher than Tai \textit{et al.}.  Furthermore, in the experiments on ResNet-34 for CIFAR-100 under CR$=$13.55, there is no significant loss in accuracy {\cred when our methods are applied.}

  \item Among the proposed methods,   LJSVD is the most outstanding one { that achieves the highest average accuracy in 8 out of 12 groups.} The RJSVD-2 is inferior to LJSVD in most cases, indicating that the left shared structure is more powerful {\cred than the right shared one}, since the only difference {\cred between  the LJSVD and RJSVD-2 is the relative position of the shared components.}

  \item {\cred RJSVD-1 outperforms RJSVD-2 in most cases, and the reason behind the phenomenon is that RJSVD-1 includes the HID layers for joint decomposition while RJSVD-2 decomposes them separately. The more layers for joint decomposition, the better the performance, which is consistent with the observation \ref{obsevation1}.}

 \item  Compared with LJSVD and RJSVD,
  Bi-JSVD$p$, $p=0.3,0.5,0.7$, can achieve not only the highest raw accuracy in most cases but also satisfying final performance. As shown in Figure \ref{line1}, the performance of Bi-JSVD$p$ {\cred increases when $p$ gets {\cbl close} to 1. However, the tendency is unsteady, indicating that the optimal $p$ is particular to the original networks and CF.} For example, it is shown that $p=1$ tends to achieve higher performance in the ResNet-50 or ResNet-34 for CIFAR-10 and CIFAR-100, while the {\cred optimal $p$} in ResNet-18 for CIFAR-10 is $0.7$. Therefore, there may not be a specific answer to the question left in the last experiment. However, empirically, {\cred one can take $p=1$, \textit{i.e.}, LJSVD, as the first choice unless the specific evaluating experiment is conducted.}

 \item Tucker Decomposition and NC\_CTD achieve the highest raw accuracies than other methods, but fail to achieve the best final performance in most cases, indicating that the final performance is not linearly correlated with the {\cred initial} accuracy but decided by the structure of the decomposed network.

  \item There is a particular phenomenon that the decomposed networks outperform the original {\cred in} ResNet-18 {\cred for} CIFAR-100. {\cred The reason for this is that decomposition followed by fine-tuning can alleviate overfitting in original networks and thus improve performance, which is consistent {\cbl with} \cite{Tai2016Convolutional}.}
 \end{enumerate}

\subsection{Evaluation on ImageNet}
\paragraph{Settings}
To further verify the effectiveness of the proposed methods, we conduct comparison experiments on a large-scale data set, ImageNet (ILSVRC-12), which contains 1.2 million images from 1000 classes. For data pre-processing, all images are normalized with $mean=[0.485, 0.456, 0.406]$ and standard deviation $std=[0.229,0.224,0.225]$, and the test images are centrally cropped to $224\times224$. For data augmentation, training images are resized to $224\times224$ using four different methods provided by Tensorflow followed by a random horizontal flip.

The ResNet-34 \cite{He2016ResNet} for ImageNet with 21.78 million of parameters is used for evaluation. To save time, the pre-trained weights are downloaded from Pytorch Model Zoo$\footnote{\url{https://download.pytorch.org/models/resnet34-333f7ec4.pth}.}$ and transferred to Tensorflow { with several epochs of fine-tuning. The network's Top-1 and Top-5 accuracies are 71.03\% and 90.25\%, respectively.}

The proposed LJSVD, RJSVD-1 and Bi-JSVD0.5 are utilized to compress the ResNet-34 following the ``pre-train$\to$decompose$\to$fine-tune'' pipeline, compared with Tai \textit{et al.}\cite{Tai2016Convolutional}, Tucker \cite{Tai2016Convolutional}  and NC\_CTD \cite{JSTSP}. Following Section \ref{Evaluation on CIFAR-10 and CIFAR-100}, $K$ for Bi-JSVD is set to 30, and conv1, conv2\_x in ResNet-34 are not decomposed. The proportion of the shared component and the independent one for NC\_CTD is set to 1:1. In the fine-tuning period, the SGD is used as the optimizer with a momentum of 0.9.
The batch size is set to 128, and the weights are regularized by L2 with a weight decay of 5e-4. Since it is very time-consuming to fine-tune ResNet-34 for ImageNet, the warmup strategy \cite{He2016ResNet,WarmupOneHour} is applied to accelerate the fine-tuning process, with which the learning rates for the first 3 epochs are set to 0.0001, 0.001 and 0.01, respectively. After the warmup stage, the compressed networks are fine-tuned for 25 epochs, with the learning rate starting from 0.1 and divided by 10 in the 6th, 11th, 16th, and 21st epochs.

\begin{table}[t]
\setlength\tabcolsep{2pt}
  \centering
    \begin{tabular}{cclccccc}
    \toprule
    \tabincell{c}{CNN\\\& acc.\\{\& \cb{FLOPs}}} & \tabincell{c}{CF\\($\times$)} & Method & \tabincell{c}{Raw\\Top-1\\acc.(\%)}& \tabincell{c}{Raw\\Top-5\\acc.(\%)} &\tabincell{c}{Top-1\\acc.\\(\%)} &\tabincell{c}{Top-5\\acc.\\(\%)}&{\cb{FLOPs}}  \\
    \midrule
    \multirow{12}[4]{*}{ \tabincell{c}{ResNet-34\\\\Top-1: 71.03\%\\Top-5: 90.14\%\\\\{\cb{7.33E9}}}} & \multirow{6}[1]{*}{10.98}
    & Tai \textit{et al.}\cite{Tai2016Convolutional} & 0.15&0.87&60.17&83.27 & {\cb{1.82E9}}\\
    &     & Tucker \cite{Kim2016Tucker}&0.15&0.79&57.64&81.40 & {\cb{1.82E9}}\\
    &     &NC\_CTD \cite{JSTSP}&\textcolor{red}{\textbf{0.38}}&\textcolor{red}{\textbf{1.60}}&58.37&82.29 & {\cb{1.98E9}}\\
    \cdashline{3-8}
     \noalign{\vskip 0.5ex}
        &     & LJSVD   &0.24&1.30&60.55&\textcolor{red}{\textbf{83.68}} & {\cb{1.90E9}}\\
        &     & RJSVD-1&0.24&1.11&60.74&83.65 & {\cb{1.92E9}}\\
        &     & Bi-JSVD0.5 &0.13&0.74&\textcolor{red}{\textbf{60.83}}&83.58 & {\cb{1.89E9}}\\
\cmidrule{2-8}
        & \multirow{6}[1]{*}{5.75} & Tai \textit{et al.}\cite{Tai2016Convolutional}&0.24&0.97&62.52&84.88 & {\cb{1.98E9}}\\
        &     & Tucker \cite{Kim2016Tucker} &\textcolor{red}{\textbf{0.84}}&\textcolor{red}{\textbf{3.09}}&58.15&81.86 & {\cb{1.97E9}}\\
        &     &NC\_CTD \cite{JSTSP}&0.39&1.67&61.12&84.26 &{\cb{ 2.29E9}}\\
 \cdashline{3-8}
  \noalign{\vskip 0.5ex}
        &     & LJSVD&0.44&1.67&63.36&85.64 & {\cb{2.15E9}}\\
        &     & RJSVD-1 &0.32&1.51&63.43&85.74 & {\cb{2.19E9}}\\
        &     & Bi-JSVD0.5&0.43&1.98&\textcolor{red}{\textbf{64.12}}&\textcolor{red}{\textbf{86.16}} &{\cb{2.14E9}}\\
 \bottomrule
    \end{tabular}
   \caption{Comparison of compressed ResNet-34 for ImageNet following the ``pre-train$\to$decompose$\to$fine-tune'' pipeline.}
   \label{results31}
\end{table}

\paragraph{Results and analysis}
Results are shown in TABLE \ref{results31}. It is shown that for ImageNet classification, the proposed methods can still outperform the state-of-the-art methods and alleviate performance degradation. The Top-1 accuracy of all the proposed methods is more than 2\% higher than NC\_CTD \cite{JSTSP} when CF=10.98, and Bi-JSVD0.5 achieves 0.52\% higher Top-5 accuracy compared with Tai \textit{et al.}\cite{Tai2016Convolutional} when CF is set to be 5.75, which further demonstrates the compatibility of the proposed algorithms and verify the effectiveness of the jointly decomposed methods for network compression.

\begin{table}[h]
\setlength\tabcolsep{4pt}
  \centering
    \begin{tabular}{cclccc}
    \toprule
    \tabincell{c}{CNN\\\& acc.\\{\cb{\& FLOPs}}} & \tabincell{l}{CF\\($\times$)} & Method & \tabincell{c}{Acc.  (\%)\\(mean$\pm$std)} &\tabincell{c}{Best\\acc. (\%)} & \tabincell{c}{\cb{FLOPs}}\\
    \midrule
    \multirow{18}[4]{*}{\tabincell{c}{ResNet-18\\\\94.80\%\\\\{\cb 11.11E8}}} & \multirow{8}[2]{*}{17.76} & Tai \textit{et al.}\cite{Tai2016Convolutional} & 92.66$\pm$0.07 &93.10 & {\cb{3.42E8}}\\
        &     & Tucker \cite{Kim2016Tucker} & 90.83$\pm$0.13 & 91.33 & {\cb{3.41E8}}\\
        &    & NC\_CTD \cite{JSTSP} &90.77$\pm$0.31   &91.56 & {\cb{3.44E8}}\\
 \cdashline{3-6}
 \noalign{\vskip 0.5ex}
        &     & LJSVD & 92.80$\pm$0.17 &  \textcolor{red}{\textbf{93.14}}& {\cb{3.45E8}}\\
        &     & RJSVD-1 & \textcolor{red}{\textbf{92.83$\pm$0.13}} & \textcolor{red}{\textbf{93.14}} & {\cb{3.50E8}}\\
        &     & RJSVD-2 & 92.74$\pm$0.13 & 92.97 & {\cb{3.45E8}}\\
        &     & Bi-JSVD0.3 & 92.70$\pm$0.19 & 93.05 & {\cb{3.45E8}} \\
        &     & Bi-JSVD0.5 & 92.78$\pm$0.11 & 93.11 & {\cb{3.44E8}}\\
        &     & Bi-JSVD0.7 & 92.70$\pm$0.09 & 93.01 & {\cb{3.45E8}}\\

\cmidrule{2-6}       & \multirow{8}[2]{*}{11.99} & Tai \textit{et al.}\cite{Tai2016Convolutional} & 92.73$\pm$0.13 & 92.94  & {\cb{3.65E8}}\\
        &     & Tucker \cite{Kim2016Tucker} & 90.94$\pm$0.15 & 91.33 & {\cb{3.63E8}}\\
        &    & NC\_CTD \cite{JSTSP} &91.22$\pm$0.13   &91.72 & {\cb{3.68E8}}\\
 \cdashline{3-6}
 \noalign{\vskip 0.5ex}
        &     & LJSVD & 93.18$\pm$0.11 & 93.46  & {\cb{3.73E8}}\\
        &     & RJSVD-1 & \textcolor{red}{\textbf{93.35$\pm$0.01}} & \textcolor{red}{\textbf{93.58}} & {\cb{3.83E8}}\\
        &     & RJSVD-2 & 93.08$\pm$0.09 & 93.43  & {\cb{3.73E8}}\\
        &     & Bi-JSVD0.3 & 93.34$\pm$0.13 & \textcolor{red}{\textbf{93.58}}  & {\cb{3.73E8}}\\
        &     & Bi-JSVD0.5 & 93.29$\pm$0.17 & 93.57 & {\cb{3.72E8}}\\
        &     & Bi-JSVD0.7 & 93.13$\pm$0.09 & 93.48 & {\cb{3.73E8}}\\

    \midrule
    \multirow{18}[3]{*}{\tabincell{c}{ResNet-34\\\\95.11\%\\\\{\cb 23.19E8}}} & \multirow{8}[2]{*}{22.07} & Tai \textit{et al.}\cite{Tai2016Convolutional} & 92.90$\pm$0.03 & 93.11  & {\cb{5.20E8}}\\
        &     & Tucker \cite{Kim2016Tucker} & 91.83$\pm$0.25 & 92.31 & {\cb{5.20E8}}\\
        &    & NC\_CTD \cite{JSTSP} & 91.03$\pm$0.13  &91.53 & {\cb{5.43E8}}\\
 \cdashline{3-6}
 \noalign{\vskip 0.5ex}
        &     & LJSVD & \textcolor{red}{\textbf{93.73$\pm$0.09}} & \textcolor{red}{\textbf{94.10}}  & {\cb{5.47E8}}\\
        &     & RJSVD-1 & 93.59$\pm$0.14 & 94.00   & {\cb{5.54E8}}\\
        &     & RJSVD-2 & 93.45$\pm$0.15 & 93.79 & {\cb{5.47E8}} \\
        &     & Bi-JSVD0.3 & 93.65$\pm$0.13 & 93.92 &{\cb{ 5.44E8}}\\
        &     & Bi-JSVD0.5 & 93.42$\pm$0.09 & 93.64  & {\cb{5.43E8}}\\
        &     & Bi-JSVD0.7 & 93.34$\pm$0.15 & 93.74 & {\cb{5.44E8}}\\

\cmidrule{2-6}       & \multirow{8}[1]{*}{13.92} & Tai \textit{et al.}\cite{Tai2016Convolutional} & 93.11$\pm$0.04 & 93.43 & {\cb{5.71E8}}\\
        &     & Tucker \cite{Kim2016Tucker} & 91.78$\pm$0.15 & 92.20  & {\cb{5.70E8}}\\
        &    & NC\_CTD \cite{JSTSP} &91.46$\pm$0.10   &91.79 & {\cb{6.16E8}}\\
 \cdashline{3-6}
 \noalign{\vskip 0.5ex}
        &     & LJSVD & \textcolor{red}{\textbf{93.85$\pm$0.19}} & \textcolor{red}{\textbf{94.23}}  & {\cb{6.27E8}}\\
        &     & RJSVD-1 & 93.62$\pm$0.15 & 93.62 & {\cb{6.42E8}}\\
        &     & RJSVD-2 & 93.49$\pm$0.14 & 93.73 & {\cb{6.27E8}}\\
        &     & Bi-JSVD0.3 & 93.36$\pm$0.07 & 93.52  & {\cb{6.22E8}}\\
        &     & Bi-JSVD0.5 & 93.60$\pm$0.09 & 93.94 & {\cb{6.24E8}}\\
        &     & Bi-JSVD0.7 & 93.57$\pm$0.06 & 93.86  & {\cb{6.22E8}}\\

\midrule
    \multirow{14}[3]{*}{\tabincell{c}{ResNet-50\\\\95.02\%\\\\{\cb 25.96E8}}} & \multirow{7}[1]{*}{6.48} & Tai \textit{et al.}\cite{Tai2016Convolutional} & 91.99$\pm$0.08 & 92.42  & {\cb{7.19E8}}\\
 \cdashline{3-6}
 \noalign{\vskip 0.5ex}
        &     & LJSVD & \textcolor{red}{\textbf{92.89$\pm$0.11}} & \textcolor{red}{\textbf{93.17}}  & {\cb{7.63E8}}\\
        &     & RJSVD-1 & 92.77$\pm$0.03 & 92.97  & {\cb{7.77E8}}\\
        &     & RJSVD-2 & 92.46$\pm$0.15 & 92.80 & {\cb{7.73E8}}\\
        &     & Bi-JSVD0.3 & 92.85$\pm$0.15 & 93.00 & {\cb{7.66E8 }}\\
        &     & Bi-JSVD0.5 & 92.83$\pm$0.18 & 93.13  & {\cb{7.66E8}}\\
        &     & Bi-JSVD0.7 & 92.64$\pm$0.20 & 92.95 & {\cb{7.61E8}}\\
\cmidrule{2-6}       & \multirow{7}[2]{*}{5.37} & Tai \textit{et al.}\cite{Tai2016Convolutional} & 92.22$\pm$0.19 & 92.55 & {\cb{7.97E8}}\\
 \cdashline{3-6}
 \noalign{\vskip 0.5ex}
        &     & LJSVD & 92.93$\pm$0.15 & 93.25  & {\cb{8.87E8}}\\
        &     & RJSVD-1 & 93.00$\pm$0.02 & 93.37  & {\cb{9.17E8}}\\
        &     & RJSVD-2 & 92.74$\pm$0.18 & 93.15  & {\cb{9.11E8}}\\
        &     & Bi-JSVD0.3 & 92.83$\pm$0.07 & 93.06  & {\cb{8.99E8}}\\
        &     & Bi-JSVD0.5 & 93.10$\pm$0.19 & 93.42  & {\cb{8.95E8}}\\
        &     & Bi-JSVD0.7 & \textcolor{red}{\textbf{93.41$\pm$0.09}} & \textcolor{red}{\textbf{93.61}} & {\cb{8.89E8}}\\
    \bottomrule
    \end{tabular}%
   \caption{Comparison of compressed ResNet trained from scratch  on CIFAR-10.}
    \label{results21}
\end{table}%

\begin{table}[h]
\setlength\tabcolsep{4pt}
  \centering
    \begin{tabular}{cclccc}
    \toprule
    \tabincell{c}{CNN\\\& acc.\\{\cb{\& FLOPs}}} & \tabincell{c}{CF\\($\times$)} & Method &\tabincell{c}{Acc.  (\%)\\(mean$\pm$std)} &\tabincell{c}{Best \\acc. (\%)} & \tabincell{c}{\cb{FLOPs}}\\

    \midrule
    \multirow{18}[4]{*}{\tabincell{c}{ResNet-18\\\\70.73\%\\\\{\cb 11.11E8}}} & \multirow{8}[2]{*}{16.61} & Tai \textit{et al.}\cite{Tai2016Convolutional} & 71.12$\pm$0.35 & 72.25& {\cb{3.42E8}} \\
 &     & Tucker \cite{Kim2016Tucker} & 69.67$\pm$0.23 & 69.88 & {\cb{3.41E8}}\\
 &    & NC\_CTD \cite{JSTSP} &70.66$\pm$0.27   &71.24 & {\cb{3.44E8}}\\
 \cdashline{3-6}
\noalign{\vskip 0.5ex}
 &     & LJSVD & 71.35$\pm$0.19 & 71.92 & {\cb{3.45E8}}\\
 &     & RJSVD-1 & \textcolor{red}{\textbf{71.59$\pm$0.39}} & \textcolor{red}{\textbf{72.48}} & {\cb{3.50E8}}\\
 &     & RJSVD-2 & 71.33$\pm$0.16 & 72.08 & {\cb{3.45E8}}\\
  &     & Bi-JSVD0.3 & 71.44$\pm$0.13  & 71.88 & {\cb{3.45E8}}\\
    &     & Bi-JSVD0.5 & 71.12$\pm$0.19 & 71.82  & {\cb{3.44E8}}\\
   &     & Bi-JSVD0.7 & 71.58$\pm$0.40 & 72.40  & {\cb{3.45E8}}\\

\cmidrule{2-6} & \multirow{8}[2]{*}{11.47} & Tai \textit{et al.}\cite{Tai2016Convolutional} & 71.63$\pm$0.25 & 72.26  & {\cb{3.65E8}}\\
       &     & Tucker \cite{Kim2016Tucker} & 69.84$\pm$0.29 & 70.92  & {\cb{3.64E8}}\\
       &    & NC\_CTD \cite{JSTSP} &70.72$\pm$0.06   &71.43  & {\cb{3.68E8}}\\
 \cdashline{3-6}
 \noalign{\vskip 0.5ex}
       &     & LJSVD & 73.13$\pm$0.31 & 73.88  & {\cb{3.73E8}}\\
         &     & RJSVD-1 & \textcolor{red}{\textbf{73.38$\pm$0.22}} & \textcolor{red}{\textbf{73.96}}  & {\cb{3.83E8}}\\
      &     & RJSVD-2 & 72.97$\pm$0.51 & 73.58  & {\cb{3.73E8}}\\
         &     & Bi-JSVD0.3 & 73.11$\pm$0.23 & 73.88  & {\cb{3.73E8}}\\
  &     & Bi-JSVD0.5 & 73.29$\pm$0.09 & 73.68  & {\cb{3.72E8}}\\
        &     & Bi-JSVD0.7 & 73.04$\pm$0.11 & 73.37  & {\cb{3.73E8}}\\

    \midrule
   \multirow{18}[3]{*}{\tabincell{c}{ResNet-34\\\\75.81\%\\\\{\cb 23.19E8}}} & \multirow{8}[2]{*}{21.11} & Tai \textit{et al.}\cite{Tai2016Convolutional} & 72.59$\pm$0.49 & 73.81  & {\cb{5.20E8}}\\
         &     & Tucker \cite{Kim2016Tucker} & 70.95$\pm$0.51 & 71.87  & {\cb{5.20E8}}\\
         &    & NC\_CTD \cite{JSTSP} & 71.52$\pm$0.03  &72.14  & {\cb{5.71E8}}\\
 \cdashline{3-6}
 \noalign{\vskip 0.5ex}
       &     & LJSVD & 73.61$\pm$0.37 & \textcolor{red}{\textbf{74.43}}  & {\cb{ 5.47E8}}\\
       &     & RJSVD-1 & \textcolor{red}{\textbf{73.66$\pm$0.25}} & 74.14 & {\cb{5.54E8}}\\
       &     & RJSVD-2 & 72.03$\pm$0.19 & 72.23  & {\cb{5.47E8}}\\
          &     & Bi-JSVD0.3 & 72.82$\pm$0.29 & 73.41  & {\cb{5.44E8}}\\
        &     & Bi-JSVD0.5 & 72.96$\pm$0.05 & 73.26  & {\cb{5.43E8}}\\
         &     & Bi-JSVD0.7 & 73.17$\pm$0.21 & 73.80  & {\cb{5.44E8}}\\

\cmidrule{2-6}  & \multirow{8}[1]{*}{13.55} & Tai \textit{et al.}\cite{Tai2016Convolutional} & 73.60$\pm$0.11 & 73.85  & {\cb{5.71E8}}\\
        &     & Tucker \cite{Kim2016Tucker} & 70.88$\pm$0.07 & 71.48  & {\cb{5.70E8}}\\
        &    & NC\_CTD \cite{JSTSP} &71.38$\pm$0.15   &72.25  & {\cb{6.16E8}}\\
 \cdashline{3-6}
 \noalign{\vskip 0.5ex}
         &     & LJSVD & \textcolor{red}{\textbf{74.89$\pm$0.19}} & \textcolor{red}{\textbf{75.29}}  & {\cb{6.27E8}}\\
        &     & RJSVD-1 & 74.38$\pm$0.31 & 74.95  & {\cb{6.42E8}}\\
        &     & RJSVD-2 & 73.75$\pm$0.12 & 74.26  & {\cb{6.27E8}}\\
        &     & Bi-JSVD0.3 & 72.84$\pm$0.17 & 73.57  & {\cb{6.22E8}}\\
        &     & Bi-JSVD0.5 & 73.13$\pm$0.29 & 73.54  & {\cb{6.25E8}}\\
         &     & Bi-JSVD0.7 & 72.50$\pm$0.35 & 73.16  & {\cb{6.22E8}}\\
\midrule
    \multirow{14}[3]{*}{\tabincell{c}{ResNet-50\\\\75.67\%\\\\{\cb 25.96E8}}} & \multirow{7}[1]{*}{6.21} & Tai \textit{et al.}\cite{Tai2016Convolutional} & 70.32$\pm$0.27 & 70.91  & {\cb{7.20E8}}\\
 \cdashline{3-6}
 \noalign{\vskip 0.5ex}
       &     & LJSVD & \textcolor{red}{\textbf{72.46$\pm$0.31}} & 73.14  & {\cb{7.63E8}}\\
        &     & RJSVD-1 & 71.93$\pm$0.41 & 72.76  & {\cb{7.78E8}}\\
         &     & RJSVD-2 & 71.40$\pm$0.20 & 71.98  & {\cb{7.74E8}}\\
          &     & Bi-JSVD0.3 & 72.04$\pm$0.83 & \textcolor{red}{\textbf{73.65}}  & {\cb{7.66E8}}\\
         &     & Bi-JSVD0.5 & 71.23$\pm$0.26 & 72.02  & {\cb{7.66E8}}\\
         &     & Bi-JSVD0.7 & 71.82$\pm$0.43 & 72.84  & {\cb{7.61E8}}\\
\cmidrule{2-6} & \multirow{7}[2]{*}{5.19} & Tai \textit{et al.}\cite{Tai2016Convolutional} & 71.35$\pm$0.38 & 72.02  & {\cb{7.97E8}}\\
 \cdashline{3-6}
 \noalign{\vskip 0.5ex}
        &     & LJSVD & 72.93$\pm$0.97 & 74.49  & {\cb{8.88E8}}\\
        &     & RJSVD-1 & 72.81$\pm$0.45 & 73.68  & {\cb{9.17E8}}\\
        &     & RJSVD-2 & 73.44$\pm$0.11 & \textcolor{red}{\textbf{74.66}}  & {\cb{9.11E8}}\\
         &     & Bi-JSVD0.3 & 72.96$\pm$0.78 & 73.88  & {\cb{9.00E8}}\\
         &     & Bi-JSVD0.5 & 72.36$\pm$0.22 & 72.95  & {\cb{8.95E8}}\\
         &     & Bi-JSVD0.7 & \textcolor{red}{\textbf{73.47$\pm$0.36}} & 74.18  & {\cb{8.90E8}}\\
    \bottomrule
    \end{tabular}%
   \caption{Comparison of compressed ResNet trained from scratch on CIFAR-100.}
    \label{results22}
\end{table}%

\subsection{Ablation Study}

\subsubsection{Train from scratch}

\paragraph{Settings}
The comparison experiments demonstrate the superior performance of the proposed methods, and we consider that there are two main reasons contributing to their superiority: the joint structures, and prior knowledge inherited from the original networks. To verify this, we further conduct ablation experiments {\cbl by} training decomposed networks from scratch, which also evaluates the ability of the proposed method in compressing CNNs in another manner. We retain the structures of the compressed networks in Section \ref{PC} and train them from scratch with {\cbl randomly initialized} parameters. Training settings such as the number of epochs, learning rate, and batch size are consistent with Section \ref{PC}.

\paragraph{Results}
The results are shown in Tables \ref{results21} and \ref{results22},
and Figure \ref{line2} shows how $p$ affects the performance of compressed networks.

\begin{figure*}[htbp]
\centering
\includegraphics[width=\linewidth]{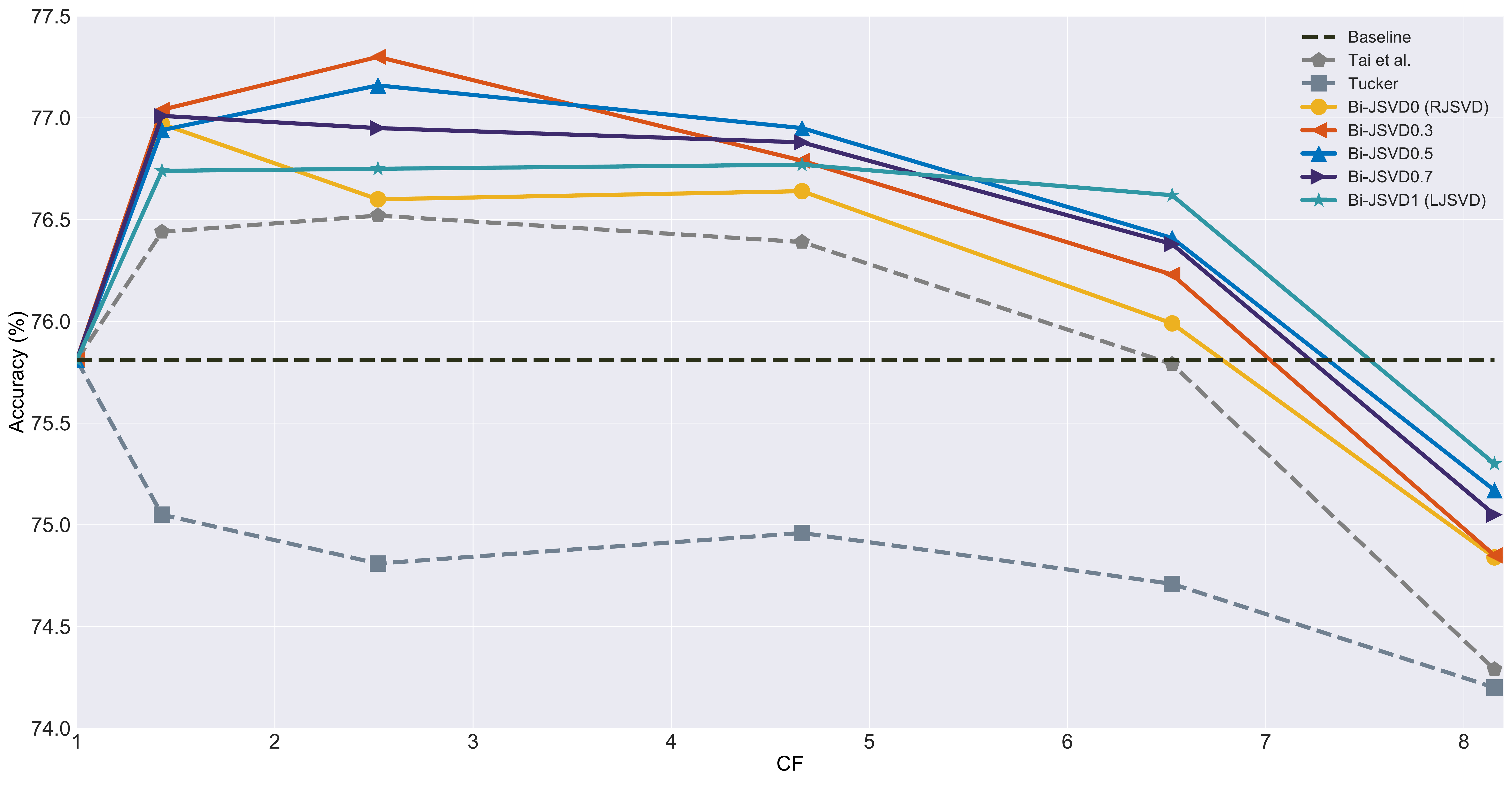}
\caption{\cb Comparison of compressed ResNet-34 for CIFAR-100 in relatively small CFs following the ``pre-train$\to$decompose$\to$fine-tune'' pipeline. HID layers are not decomposed. The results are the average accuracy of three times repeated experiments.}
\label{CF}
\end{figure*}

\begin{figure*}[htbp]
\centering
\includegraphics[width=\linewidth]{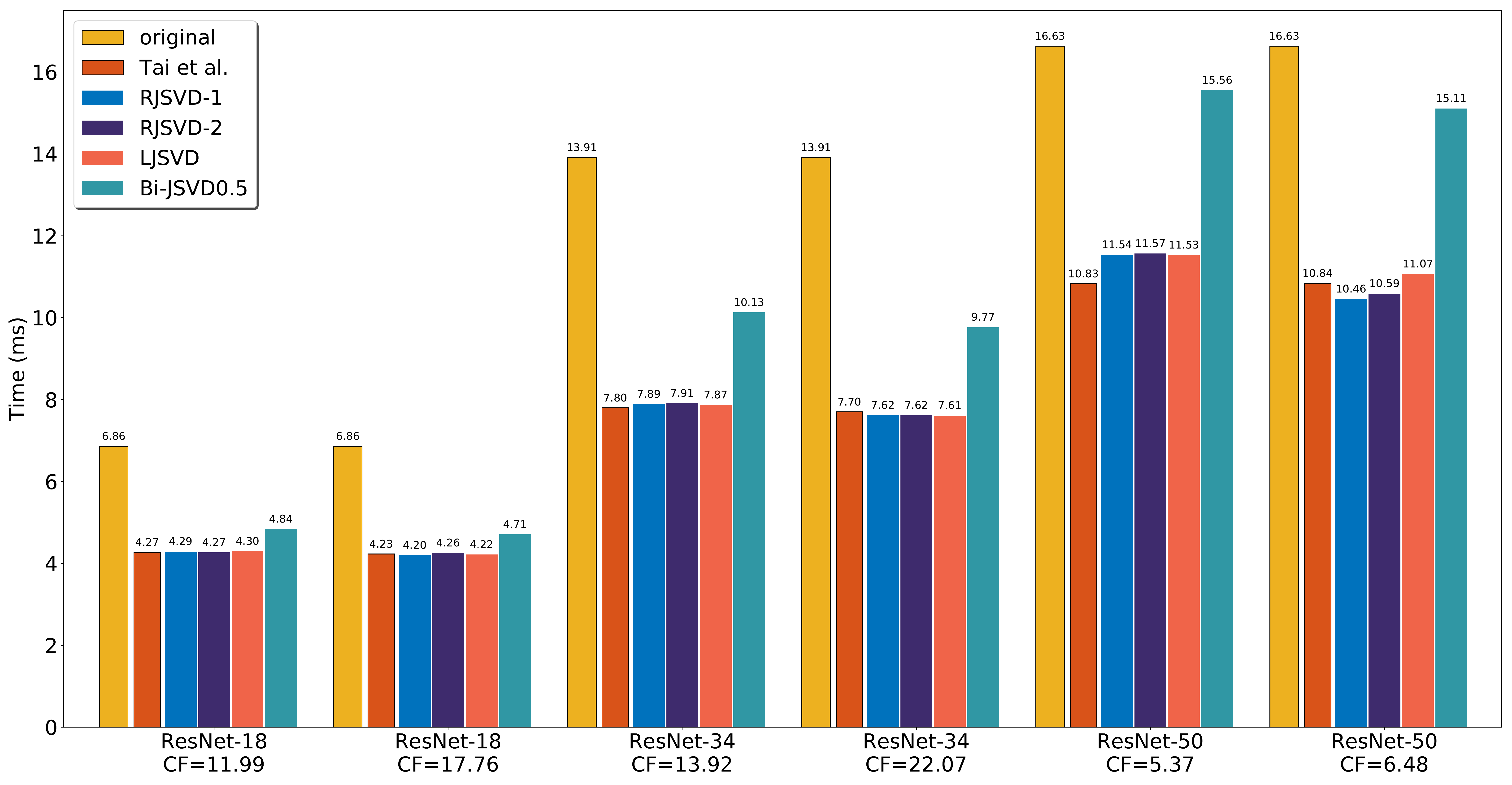}
\caption{\cb{Realistic acceleration for the forward procedure on CIFAR-10 (evaluated on an Intel E5-2690@2.6GHz CPU; repeated 50,000 times to obtain the average time consumption). Accuracy of the corresponding networks are shown in the Table \ref{results11}.}}
\label{fig:accelerate}
\end{figure*}

\paragraph{Analysis}
Not surprisingly,
the proposed methods outperform the baseline methods, demonstrating the power of the proposed joint structures {\cbl in} alleviating performance degradation, {\cbl and} conclusions similar to the previous subsection can still be summarized.
However,
there are some outcomes different from the previous results as follows:
\begin{enumerate}[(i)]
\item {Compared with results in Table \ref{results11} and \ref{results12},
the accuracy of compressed networks trained from scratch} drops in different degrees, implying that factorized weights inheriting the prior knowledge from the original networks can give a more appropriate initialization to help CNNs settle at a better local minimum. Furthermore, the drops {of} the proposed methods are much smaller than the baseline approaches in most cases. For example, LJSVD {\cbl from scratch} outperforms Tai \textit{et al.} with 2.14\% higher accuracy on ResNet-50 for CIFAR-100 when $\textrm{CR}=6.21$, while it is 0.68\% {\cbl with pre-train weights}. {\cbl This observation demonstrates} that the proposed methods are more initialization-independent, {\cbl and thus} the joint structure is the main factor in improving performance.

\item Different from  Figure \ref{line1},
the curves in  Figure \ref{line2} are  more messy and have no  definite trends, indicating that compressed networks {\cbl with pre-train weights} are trained following specific patterns decided by the original networks, whereas the ones trained from scratch are in random.
\end{enumerate}

\subsubsection{Varying CF}
\paragraph{Settings}
In this section, we verify the performance of the proposed methods under different CFs on ResNet-34 for CIFAR-100. Different from Section \ref{PC} that compresses networks with large CFs and decomposes the HID layers, we keep the HID layers uncompressed and compare the proposed methods with Tai \textit{et al.}\cite{Tai2016Convolutional}  and Tucker \cite{Kim2016Tucker} under relatively small CFs ranking from 1.43 to 8.15. Other settings, such as fine-tuning schedules and repeated times, are the same as  those in Section \ref{PC}.

\paragraph{Results and Analysis}
The results are illustrated in Figure \ref{CF}. It is shown that all the proposed methods outperform the counterparts under small CFs. Moreover, the proposed methods can effectively alleviate the over-fitting problem under small CFs and thus achieve higher accuracy than the original network, which is also observed in Table \ref{results22}. Furthermore, Bi-JSVD$p$ for $p \in [0.3,0.5,0.7]$ are superior to RJSVD and LJSVD when CF is less than 5, while LJSVD achieves the best performance in other scenarios, followed by Bi-JSVD and then RJSVD. We believe that the reasons for different performance of the proposed methods are as follows:
\begin{enumerate}[(i)]
\item - Why can LJSVD outperform RJSVD?\\
- The only difference between LJSVD and RJSVD is the relative position of the shared sub-convolutional layers obtained by the joint decomposition, as shown in Figure \ref{fig:net}. {\cred Although the shared sub-tensors are necessary to approximate the original weight tensor,} they lack specificity for each layer because of their shareability, and thus may yield vague feature maps. In RJSVD, the shared sub-convolutional layers come after the independent ones, {\cbl making} feature maps passed to the next {\cbl blocks} vague. While in LJSVD, {\cbl the contrary is the case, and thus} explicit features can be captured by following layers, which improves the performance of decomposed networks {\cred and makes LJSVD superior to RJSVD.}

\item - Why can Bi-JSVD achieve good performance?\\
- As shown in Figure \ref{fig:net}, Bi-JSVD$p$ ($p\notin$ \{0,1\}) yields dual paths, which in fact widens the networks.
{\cred Similar to the Inception Module in GoogLeNet \cite{GoogleNet}, the dual paths can enrich  the information carried by feature maps in each layer, {\cbl thus improving} the performance of BI-JSVD.}

\item - Why does the impact of $p$ in Bi-JSVD seem irregular?\\
- As discussed above, LJSVD has its advantages compared with RJSVD. Therefore, one may expect better performance with a larger $p$ in Bi-JSVD$p$ {\cbl since Bi-JSVD$0$ is equivalent to RJSVD and Bi-JSVD$1$ to LJSVD}. However, the extreme $p=1$ ({\cred single-branch LJSVD) would} eliminate the advantage of dual paths in Bi-JSVD. Therefore, there is a trade-off between the dual paths and left-shared structure, and $p$ is the key for the balance. However, the optimal $p$ depends on baseline networks and CFs, {\cred making} the impact of $p$ irregular. Since there is no uniform and closed-form solution to $p$, AutoML methods \cite{junhao20} can be utilized to find optimal $p$  for specific scenarios in future work.
\end{enumerate}

\subsection{Realistic Acceleration}
\paragraph{Settings}
As mentioned in Section \ref{sec:methodology}, the complexity of compressed convolutional layers produced by the proposed methods is less than the original ones, thus the forward inference can be accelerated. However, there is a wide gap between theoretical and realistic acceleration, which is restricted by the IO delay, buffer switch, efficiency of BLAS libraries \cite{GeometricFilterPruning} and some indecomposable layers such as batch normalization and pooling. Therefore, to evaluate the realistic acceleration fairly, we measure the real forward time of the networks compressed {\cb by Tai \textit{et al.}\cite{Tai2016Convolutional} based on the matrix decomposition,} and {\cb the proposed} LJSVD, RJSVD-1, RJSVD-2 and Bi-JSVD0.5 in TABLE \ref{results11}.
Each method is evaluated on an Intel E5-2690@2.6GHz CPU, and the measurement is repeated 50,000 times to calculate the average time cost for one image by setting the batch size to 1.

\paragraph{Results and analysis} As shown in Figure \ref{fig:accelerate},
all the networks are accelerated {\cbl without} support from {\cred customized hardware}. The proposed LJSVD, RJSVD-1 and RJSVD-2 can {\cbl significantly accelerate networks as Tai \textit{et al.}\cite{Tai2016Convolutional} but with much higher accuracy}, especially on ResNet-18 and ResNet-34. Since ResNet-50 is built with BottleNecks consisting of $1\times1$ convolutional layers, the acceleration on ResNet-50 is less obvious than on ResNet-18 and ResNet-34. {\cbl Moreover, Bi-JSVD is not as outstanding in acceleration as LJSVD and RJSVD, since the dual paths shown in Figure \ref{fig:net} cause more processing time and IO delay}. Furthermore, it is shown that {\cbl a relatively significant gap in FLOPs brings little difference in inference time,} {\cbl indicating that  IO delay and buffer switch mentioned above consume  a large proportion of inference time.}

\subsection{Discussion of the proposed methods}
The afore-presented experiments have demonstrated the effectiveness of the proposed RJSVD, LJSVD and Bi-JSVD. However, they have their own cons and pros:
\begin{enumerate}[(i)]
\item
Bi-JSVD widens the networks and thus can provide satisfactory results.
{\cbl However, it requires} efforts to find the optimal $p$ for specific baseline models, {\cbl costs more inference time than RJSVD and LJSVD, and consumes about twice the memory cache storing activations due to the dual paths}.

\item Both RJSVD and LJSVD are faster than Bi-JSVD$p$ ($p \notin \{0,1\}$),
but RJSVD is inferior to LJSVD because the shared sub-convolution layers {\cred in RJSVD} are located before the independent ones. However, RJSVD is able to compress the HID layers, while LJSVD and Bi-JSVD fail to do so. As shown in Table IV--VII, RJSVD-1 that takes into account HID layers for joint decomposition can also provide competitive results.

\end{enumerate}
Therefore, the three proposed variants are suitable for different scenarios. Bi-JSVD with a procedure verifying $p$ is suitable for specific scenarios that require not only large CFs but also superior performance. For RJSVD, it would be a good choice if there are many HID layers to be decomposed jointly, while LJSVD is appropriate when there are requirements for fast inference and satisfactory performance.

\section{Conclusion And Future Work}
\label{sec:conclusion}
In this paper, inspired by the fact that there are repeated modules among CNNs,
we propose to compress networks jointly to alleviate performance degradation. Three joint matrix decomposition algorithms, RJSVD, LJSVD and {\cbl their generalization}, Bi-JSVD, are introduced. Extensive experiments on  CIFAR-10, CIFAR-100 and the larger scale ImageNet show that our methods can achieve better compression results {\cred compared} with the state-of-the-art matrix or tensor-based methods, and the speed acceleration in the forward inference period is verified as well.

In future work, we plan to extend the joint {\cbl compression} method to other areas such as network structured pruning to further improve the acceleration performance of CNNs, and use AutoML techniques such as Neural Architecture Search \cite{NAS2020ICLR}, Reinforcement Learning \cite{DeepQLearning} and Evolutionary Algorithms \cite{NSGA-II,MOEA/D} to estimate the {\cred optimal} $p$ for Bi-JSVD as well as target ranks {\cred to} compress networks adaptively.

\section*{Acknowledgement}
The work described in this paper was supported in part by the Guangdong Basic and Applied Basic Research Foundation under Grant 2021A1515011706, in part by National Science Fund for Distinguished Young Scholars under Grant 61925108, in part by the Foundation of Shenzhen under Grant JCYJ20190808122005605, in part by Shenzhen Stabilization Support Grant 20200809153412001, and in part  by the Open Research Fund from the Guangdong Provincial Key Laboratory of Big Data Computing, The Chinese University of Hong Kong, Shenzhen, under Grant No. B10120210117-OF07.


\begin{thebibliography}{10}
\providecommand{\url}[1]{#1}
\csname url@samestyle\endcsname
\providecommand{\newblock}{\relax}
\providecommand{\bibinfo}[2]{#2}
\providecommand{\BIBentrySTDinterwordspacing}{\spaceskip=0pt\relax}
\providecommand{\BIBentryALTinterwordstretchfactor}{4}
\providecommand{\BIBentryALTinterwordspacing}{\spaceskip=\fontdimen2\font plus
\BIBentryALTinterwordstretchfactor\fontdimen3\font minus
  \fontdimen4\font\relax}
\providecommand{\BIBforeignlanguage}[2]{{%
\expandafter\ifx\csname l@#1\endcsname\relax
\typeout{** WARNING: IEEEtran.bst: No hyphenation pattern has been}%
\typeout{** loaded for the language `#1'. Using the pattern for}%
\typeout{** the default language instead.}%
\else
\language=\csname l@#1\endcsname
\fi
#2}}
\providecommand{\BIBdecl}{\relax}
\BIBdecl

\bibitem{DBLP:journals/ijon/GuanLXLG20}
J.~Guan, R.~Lai, A.~Xiong, Z.~Liu, and L.~Gu, ``Fixed pattern noise reduction
  for infrared images based on cascade residual attention {CNN},''
  \emph{Neurocomputing}, vol. 377, pp. 301--313, 2020.

\bibitem{DBLP:conf/aaai/TaherkhaniKN19}
F.~Taherkhani, H.~Kazemi, and N.~M. Nasrabadi, ``Matrix completion for
  graph-based deep semi-supervised learning,'' in \emph{AAAI}, 2019, pp.
  5058--5065.

\bibitem{Girshick2014ObjectDetection}
R.~B. Girshick, J.~Donahue, T.~Darrell, and J.~Malik, ``Rich feature
  hierarchies for accurate object detection and semantic segmentation,'' in
  \emph{{CVPR}}, 2014, pp. 580--587.

\bibitem{EfficientNet}
M.~Tan and Q.~V. Le, ``Efficientnet: Rethinking model scaling for convolutional
  neural networks,'' in \emph{{ICML}}, vol.~97, 2019, pp. 6105--6114.

\bibitem{VGG}
K.~Simonyan and A.~Zisserman, ``Very deep convolutional networks for
  large-scale image recognition,'' in \emph{{ICLR}}, 2015.

\bibitem{He2016ResNet}
K.~He, X.~Zhang, S.~Ren, and J.~Sun, ``Deep residual learning for image
  recognition,'' in \emph{{CVPR}}, 2016, pp. 770--778.

\bibitem{GoogleNet}
C.~Szegedy, W.~Liu, Y.~Jia, P.~Sermanet, S.~E. Reed, D.~Anguelov, D.~Erhan,
  V.~Vanhoucke, and A.~Rabinovich, ``Going deeper with convolutions,'' in
  \emph{CVPR}, 2015, pp. 1--9.

\bibitem{Kim2016Tucker}
Y.~Kim, E.~Park, S.~Yoo, T.~Choi, L.~Yang, and D.~Shin, ``Compression of deep
  convolutional neural networks for fast and low power mobile applications,''
  in \emph{{ICLR}}, 2016.

\bibitem{SqueezeNet}
F.~N. Iandola, M.~W. Moskewicz, K.~Ashraf, S.~Han, W.~J. Dally, and K.~Keutzer,
  ``Squeezenet: Alexnet-level accuracy with 50x fewer parameters and $<$0.5mb
  model size,'' \emph{arXiv:1602.07360}, 2016.

\bibitem{Denil2013Predicting}
M.~Denil, B.~Shakibi, L.~Dinh, M.~Ranzato, and N.~de~Freitas, ``Predicting
  parameters in deep learning,'' in \emph{{NeurIPS}}, 2013, pp. 2148--2156.

\bibitem{Denton2014Exploiting}
E.~L. Denton, W.~Zaremba, J.~Bruna, Y.~LeCun, and R.~Fergus, ``Exploiting
  linear structure within convolutional networks for efficient evaluation,'' in
  \emph{{NeurIPS}}, 2014, pp. 1269--1277.

\bibitem{Jaderberg2014Speeding}
M.~Jaderberg, A.~Vedaldi, and A.~Zisserman, ``Speeding up convolutional neural
  networks with low rank expansions,'' in \emph{{BMVC}}, 2014.

\bibitem{Tai2016Convolutional}
C.~Tai, T.~Xiao, X.~Wang, and W.~E, ``Convolutional neural networks with
  low-rank regularization,'' in \emph{{ICLR} (Poster)}, 2016.

\bibitem{Novikov2015TT}
A.~Novikov, D.~Podoprikhin, A.~Osokin, and D.~P. Vetrov, ``Tensorizing neural
  networks,'' in \emph{{NeurIPS}}, 2015, pp. 442--450.

\bibitem{CPICLR}
V.~Lebedev, Y.~Ganin, M.~Rakhuba, I.~V. Oseledets, and V.~S. Lempitsky,
  ``Speeding-up convolutional neural networks using fine-tuned
  cp-decomposition,'' in \emph{{ICLR} (Poster)}, 2015.

\bibitem{Garipov2016TTalike}
T.~Garipov, D.~Podoprikhin, A.~Novikov, and D.~P. Vetrov, ``Ultimate
  tensorization: compressing convolutional and {FC} layers alike,''
  \emph{arXiv:1611.03214}, 2016.

\bibitem{Wang2018TensorRing}
W.~Wang, Y.~Sun, B.~Eriksson, W.~Wang, and V.~Aggarwal, ``Wide compression:
  Tensor ring nets,'' in \emph{{CVPR}}, 2018, pp. 9329--9338.

\bibitem{Yu2017LowRankSparse}
X.~Yu, T.~Liu, X.~Wang, and D.~Tao, ``On compressing deep models by low rank
  and sparse decomposition,'' in \emph{{CVPR}}, 2017, pp. 67--76.

\bibitem{junhao20}
J.~Huang, W.~Sun, and L.~Huang, ``Deep neural networks compression learning
  based on multiobjective evolutionary algorithms,'' \emph{Neurocomputing},
  vol. 378, pp. 260--269, 2020.

\bibitem{JSTSP}
W.~Sun, S.~Chen, L.~Huang, H.~C. So, and M.~Xie, ``Deep convolutional neural
  network compression via coupled tensor decomposition,'' \emph{{IEEE} J. Sel.
  Top. Signal Process.}, vol.~15, no.~3, pp. 603--616, 2021.

\bibitem{TensorTrainDecompositon}
I.~V. Oseledets, ``Tensor-train decomposition,'' \emph{{SIAM} J. Sci. Comput.},
  vol.~33, no.~5, pp. 2295--2317, 2011.

\bibitem{lecun1990optimal}
Y.~LeCun, J.~S. Denker, and S.~A. Solla, ``Optimal brain damage,'' in
  \emph{NeurIPS}, 1989, pp. 598--605.

\bibitem{han2015learning}
S.~Han, J.~Pool, J.~Tran, and W.~J. Dally, ``Learning both weights and
  connections for efficient neural network,'' in \emph{NeurIPS}, 2015, pp.
  1135--1143.

\bibitem{LotteryTicket}
J.~Frankle and M.~Carbin, ``The lottery ticket hypothesis: Finding sparse,
  trainable neural networks,'' in \emph{ICLR}, 2019.

\bibitem{Chen2021MOEA}
J.~Chen, Y.~Xu, W.~Sun, and L.~Huang, ``Joint sparse neural network compression
  via multi-application multi-objective optimization,'' \emph{Appl. Intell.},
  vol.~51, no.~11, pp. 7837--7854, 2021.

\bibitem{PPCA}
J.~Zhou, H.~Qi, Y.~Chen, and H.~Wang, ``Progressive principle component
  analysis for compressing deep convolutional neural networks,''
  \emph{Neurocomputing}, vol. 440, pp. 197--206, 2021.

\bibitem{GeometricFilterPruning}
Y.~He, P.~Liu, Z.~Wang, Z.~Hu, and Y.~Yang, ``Filter pruning via geometric
  median for deep convolutional neural networks acceleration,'' in
  \emph{{CVPR}}, 2019, pp. 4340--4349.

\bibitem{JacobInteger}
B.~Jacob, S.~Kligys, B.~Chen, M.~Zhu, M.~Tang, A.~G. Howard, H.~Adam, and
  D.~Kalenichenko, ``Quantization and training of neural networks for efficient
  integer-arithmetic-only inference,'' in \emph{CVPR}, 2018, pp. 2704--2713.

\bibitem{XNOR}
M.~Rastegari, V.~Ordonez, J.~Redmon, and A.~Farhadi, ``Xnor-net: Imagenet
  classification using binary convolutional neural networks,'' in \emph{ECCV},
  ser. Lecture Notes in Computer Science, B.~Leibe, J.~Matas, N.~Sebe, and
  M.~Welling, Eds., vol. 9908, 2016, pp. 525--542.

\bibitem{HanDeepCompression}
S.~Han, H.~Mao, and W.~J. Dally, ``Deep compression: Compressing deep neural
  network with pruning, trained quantization and huffman coding,'' in
  \emph{ICLR}, Y.~Bengio and Y.~LeCun, Eds., 2016.

\bibitem{WangPostQuan}
P.~Wang, Q.~Chen, X.~He, and J.~Cheng, ``Towards accurate post-training network
  quantization via bit-split and stitching,'' in \emph{ICML}, vol. 119, 2020,
  pp. 9847--9856.

\bibitem{hinton2015distilling}
G.~Hinton, O.~Vinyals, and J.~Dean, ``Distilling the knowledge in a neural
  network,'' in \emph{NeurIPS Deep Learning Workshop}, 2015.

\bibitem{Romero2015KD}
A.~Romero, N.~Ballas, S.~E. Kahou, A.~Chassang, C.~Gatta, and Y.~Bengio,
  ``Fitnets: Hints for thin deep nets,'' in \emph{ICLR}, 2015.

\bibitem{Chen2016KD}
T.~Chen, I.~J. Goodfellow, and J.~Shlens, ``Net2net: Accelerating learning via
  knowledge transfer,'' in \emph{ICLR}, 2016.

\bibitem{Li2020KD}
T.~Li, J.~Li, Z.~Liu, and C.~Zhang, ``Few sample knowledge distillation for
  efficient network compression,'' in \emph{CVPR}, 2020, pp. 14\,627--14\,635.

\bibitem{Srinivas2018KD}
S.~Srinivas and F.~Fleuret, ``Knowledge transfer with jacobian matching,'' in
  \emph{ICML}, vol.~80, 2018, pp. 4730--4738.

\bibitem{ICA}
Q.~V. Le, A.~Karpenko, J.~Ngiam, and A.~Y. Ng, ``{ICA} with reconstruction cost
  for efficient overcomplete feature learning,'' in \emph{{NeurIPS}}, 2011, pp.
  1017--1025.

\bibitem{Sun2017SVDTDNN}
M.~Sun, D.~Snyder, Y.~Gao, V.~K. Nagaraja, M.~Rodehorst, S.~Panchapagesan,
  N.~Strom, S.~Matsoukas, and S.~Vitaladevuni, ``Compressed time delay neural
  network for small-footprint keyword spotting,'' in \emph{{INTERSPEECH}},
  2017, pp. 3607--3611.

\bibitem{SparseLowRank}
S.~Swaminathan, D.~Garg, R.~Kannan, and F.~Andr{\`{e}}s, ``Sparse low rank
  factorization for deep neural network compression,'' \emph{Neurocomputing},
  vol. 398, pp. 185--196, 2020.

\bibitem{Decoupling}
J.~Guo, Y.~Li, W.~Lin, Y.~Chen, and J.~Li, ``Network decoupling: From regular
  to depthwise separable convolutions,'' in \emph{{BMVC}}, 2018, p. 248.

\bibitem{MobileNets}
A.~G. Howard, M.~Zhu, B.~Chen, D.~Kalenichenko, W.~Wang, T.~Weyand,
  M.~Andreetto, and H.~Adam, ``Mobilenets: Efficient convolutional neural
  networks for mobile vision applications,'' \emph{arXiv:1704.04861}, 2017.

\bibitem{wu2020hybrid}
B.~Wu, D.~Wang, G.~Zhao, L.~Deng, and G.~Li, ``Hybrid tensor decomposition in
  neural network compression,'' \emph{Neural Netw}, vol. 132, pp. 309--320,
  2020.

\bibitem{Cifar-10Dataset}
A.~Krizhevsky and G.~Hinton, ``Learning multiple layers of features from tiny
  images,'' Tech. Rep., 2009.

\bibitem{WarmupOneHour}
P.~Goyal, P.~Doll{\'{a}}r, R.~B. Girshick, P.~Noordhuis, L.~Wesolowski,
  A.~Kyrola, A.~Tulloch, Y.~Jia, and K.~He, ``Accurate, large minibatch {SGD:}
  training imagenet in 1 hour,'' \emph{arXiv:1706.02677}, 2017.

\bibitem{NAS2020ICLR}
X.~Dong and Y.~Yang, ``Nas-bench-201: Extending the scope of reproducible
  neural architecture search,'' in \emph{{ICLR}}, 2020.

\bibitem{DeepQLearning}
V.~Mnih, K.~Kavukcuoglu, D.~Silver, A.~A. Rusu, J.~Veness, M.~G. Bellemare,
  A.~Graves, M.~A. Riedmiller, A.~Fidjeland, G.~Ostrovski, S.~Petersen,
  C.~Beattie, A.~Sadik, I.~Antonoglou, H.~King, D.~Kumaran, D.~Wierstra,
  S.~Legg, and D.~Hassabis, ``Human-level control through deep reinforcement
  learning,'' \emph{Nat.}, vol. 518, no. 7540, pp. 529--533, 2015.

\bibitem{NSGA-II}
K.~Deb, S.~Agrawal, A.~Pratap, and T.~Meyarivan, ``A fast and elitist
  multiobjective genetic algorithm: {NSGA-II},'' \emph{{IEEE} Trans. Evol.
  Comput.}, vol.~6, no.~2, pp. 182--197, 2002.

\bibitem{MOEA/D}
Q.~Zhang and H.~Li, ``Moea/d: A multiobjective evolutionary algorithm based on
  decomposition,'' \emph{{IEEE} Trans. Evol. Comput.}, vol.~11, no.~6, pp.
  712--731, 2007.

\end{thebibliography}

\bibliographystyle{IEEEtran}
\end{document}